\documentclass[sigconf]{acmart}

\AtBeginDocument{%
  \providecommand\BibTeX{{%
    \normalfont B\kern-0.5em{\scshape i\kern-0.25em b}\kern-0.8em\TeX}}}

\copyrightyear{2025}
\acmYear{2025}
\acmConference[GECCO '25]{Genetic and Evolutionary Computation Conference}{July 14--18, 2025}{Malaga, Spain}
\acmBooktitle{Genetic and Evolutionary Computation Conference (GECCO '25), July 14--18, 2025, Malaga, Spain}\acmDOI{10.1145/3712256.3726368}
\acmISBN{979-8-4007-1465-8/2025/07}
\setcopyright{cc}
\setcctype{by-nc}

\usepackage{booktabs}
\usepackage{tabularx}
\usepackage{csquotes}
\begin{document}

\title{Controller Distillation Reduces Fragile Brain-Body Co-Adaptation and Enables Migrations in MAP-Elites
}

\author{Alican Mertan}
\email{alican.mertan@uvm.edu}
\affiliation{%
  \institution{Neurobotics Lab \\ University of Vermont}
  \city{Burlington}
  \state{VT}
  \country{USA}
}

\author{Nick Cheney}
\email{ncheney@uvm.edu}
\affiliation{%
  \institution{Neurobotics Lab \\ University of Vermont}
  \city{Burlington}
  \state{VT}
  \country{USA}}


\begin{abstract}
  Brain-body co-optimization suffers from fragile co-adaptation where brains become over-specialized for particular bodies, hindering their ability to transfer well to others. Evolutionary algorithms tend to discard such low-performing solutions, eliminating promising morphologies. Previous work considered applying MAP-Elites, where niche descriptors are based on morphological features, to promote better search over morphology space. In this work, we show that this approach still suffers from fragile co-adaptation: where a core mechanism of MAP-Elites, creating stepping stones through solutions that migrate from one niche to another, is disrupted.  We suggest that this disruption occurs because the body mutations that move an offspring to a new morphological niche break the robots' fragile brain-body co-adaptation and thus significantly decrease the performance of those potential solutions -- reducing their likelihood of outcompeting an existing elite in that new niche. We utilize a technique, we call \textit{Pollination}, that periodically replaces the controllers of certain solutions with a distilled controller with better generalization across morphologies to reduce fragile brain-body co-adaptation and thus promote MAP-Elites migrations. Pollination increases the success of body mutations and the number of migrations, resulting in better quality-diversity metrics. We believe we develop important insights that could apply to other domains where MAP-Elites is used.

\end{abstract}

\ccsdesc[500]{Computer systems organization~Evolutionary robotics}
\ccsdesc[500]{Theory of computation~Evolutionary algorithms}

\keywords{evolutionary robotics, soft robotics, brain-body co-optimization, Quality-Diversity, MAP-Elites}

\begin{teaserfigure}
  \includegraphics[width=\textwidth]{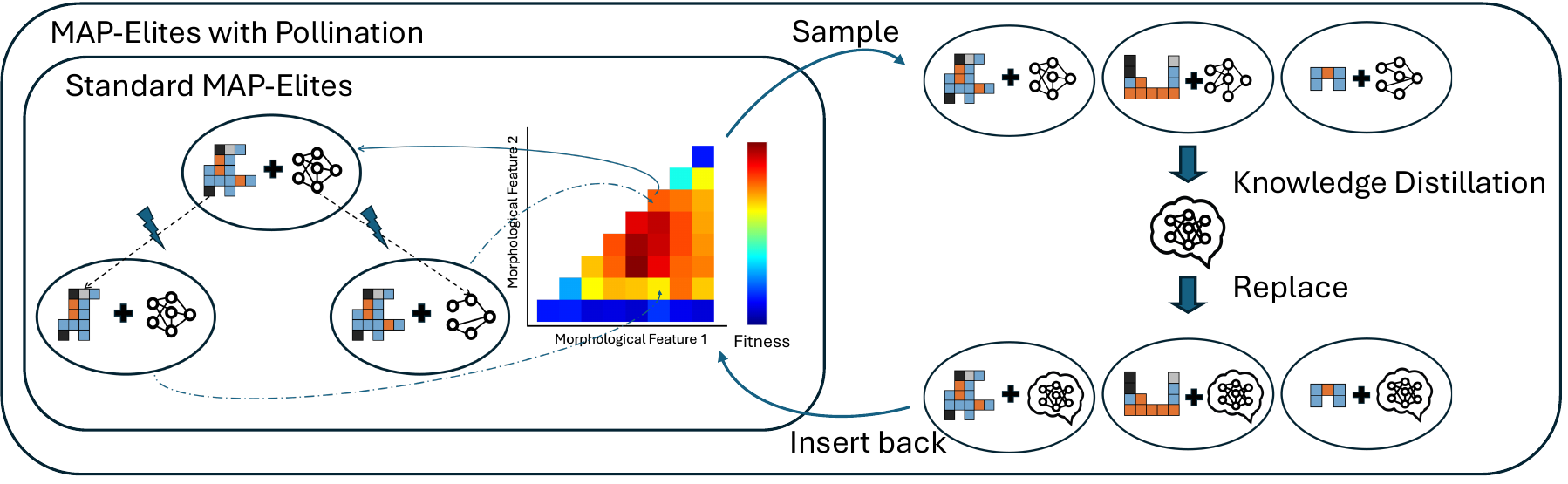} 
  \caption{Simplified schematic diagram of standard MAP-Elites algorithm applied to brain-body co-optimization and the proposed pollination extension. The standard approach suffers from fragile co-adaptation, where controllers become too specialized for particular morphologies and do not transfer well to others.  This results in a low number of migrations between niches and hinders MAP-Elites' ability to discover stepping stones. Periodically applying knowledge distillation to create multi-morphology controllers with better zero-shot transfer capabilities and injecting these controllers back into the population encourages more migrations and improves the performance of the MAP-Elites on brain-body co-optimization.}
  \Description{Teaser} 
  \label{fig:teaser}
  \vspace{1em}
\end{teaserfigure}

\maketitle

\section{Introduction}

From Karl Sims' virtual creatures~\cite{sims_evolving_1994} to robots made of living cells~\cite{kriegman_scalable_2020}, it has always been an interesting challenge to optimize agents in their entirety for a task in an automated fashion. In recent years, we have seen a surge of interest in brain-body co-optimization, where agents are conceptualized as a combination of morphology (often modeled by rigid links and joints, or volumetrically actuated voxels, in simulation) and controller (often modeled by neural networks)~\cite{mertan_modular_2023,nordmoen2021map,mertan2024investigating,bhatia2021evolution,pigozzi_evolving_2022,cheney_scalable_2018,cheney_difficulty_2016,medvet_biodiversity_2021,tanaka_co-evolving_2022,thomson_understanding_2024,marzougui_comparative_2022,gupta_embodied_2021}. Optimizing both of these components autonomously yields complete agents that are well suited to their task, often outperforming hand-designed agents~\cite{bhatia2021evolution}.

While such a framework, where agents consist of two interdependent components of morphology and controller, is intuitive, it presents a unique optimization problem with its own challenges. This has been investigated in the context of evolutionary optimization~\cite{cheney_difficulty_2016,mertan2024investigating}, and the main challenge has been identified as fragile co-adaptation: Controllers over-specialize for particular bodies and do not transfer well to others. This decreases the success of body mutations as they result in offspring with significantly decreased fitness~\cite{mertan_modular_2023}. Importantly, this holds true even when the new morphology has better potential (higher fitness if both old and new morphologies are paired with an optimal controller). This initial decrease in fitness results in poor search over the morphology space as evolutionary algorithms tend to discard low-performing solutions, eventually resulting in premature convergence of morphology. 

Previous work has reported the successful application of the MAP-Elites algorithm to the problem of brain-body co-optimization \cite{nordmoen2021map}. The use of quality-diversity algorithms such as MAP-Elites~\cite{mouret_illuminating_2015} seems intuitive to alleviate the fragile co-adaptation -- if the search over the morphology space is failing, one could promote diversity in that space by choosing the niche descriptors based on morphological features to enable a better search. However, on closer inspection, an issue arises.  A core mechanism of the MAP-Elites algorithm, the creation of often-unintuitive stepping stones~\cite{lehman2008exploiting,lehman_abandoning_2011} through solutions migrating from one niche to another~\cite{mouret_illuminating_2015} relies on the transfer of offspring from one (morphological) niche to another.  This does not play well with the phenomenon of fragile co-adaptation in brain-body co-optimization: if an offspring with a morphological mutation has significantly lowered fitness, it has a very low probability of outcompeting an existing elite in another niche -- even when it has significantly better long-term potential in that niche. To what extent are we hindering this quality-diversity algorithm if it cannot create stepping stones through migrations between niches?

To reconcile the understanding of fragile co-adaptation and successful application of MAP-Elites reported in previous work, we investigate the dynamics of MAP-Elites on the problem of brain-body co-optimization. 
Sect.~\ref{sect:investigation} discusses our application of the MAP-Elites algorithm, following its standard implementation in brain-body co-optimization, where niches are determined based on morphological features. In particular, we examine the dynamics of migrations and try to empirically determine their usefulness. Our findings suggest that stepping stones do not contribute to the performance of the best solutions found in the case of traditional applications of MAP-Elites to brain-body co-optimization. 
To promote migrations and improve MAP-Elites, we describe a method that we call \textit{Pollination} in Sect.~\ref{sect:pollination}. This method consists of periodically distilling a controller from the current archive, which has been shown to create controllers that can generalize well to different morphologies~\cite{mertan2024towards}, and injecting this generalist controller back into the archive to promote migrations. Our results demonstrate that this procedure improves the quality-diversity metrics. 
Lastly, in Sect.~\ref{sect:discussion}, we discuss the shortcomings of our approach, future work directions, and how our proposed approach connects to broader themes in quality-diversity algorithms and evolutionary computation. 

Overall, the main contributions of this work are to:
\begin{itemize}
    \item show that in the standard MAP-Elites for brain-body co-optimization, the number of migrations decreases rapidly and migrations do not contribute to the best solution found at the end of a run (Fig.~\ref{fig:standard-results},~\ref{fig:standard-results-details},~\ref{fig:compare-std-control}).
    \item show that injecting distilled controllers into the population that better generalize across morphologies (\textquote{pollination}) helps to alleviate fragile co-adaptation and results in more successful body mutations and more migrations (Fig.~\ref{fig:body-mutation-fitness-change-detailed},~\ref{fig:compare-std-pollination-migration}).
    \item demonstrate that the pollination-induced migrations strongly correlate with quality-diversity metrics and that MAP-Elites with the proposed extension outperforms the baseline MAP-Elites algorithm for brain-body co-optimization (Fig.~\ref{fig:pollination-performance-vs-migration},~\ref{fig:compare-std-pollination},~\ref{fig:behavior-collages}).
\end{itemize}

\section{Laying out the Problem}\label{sect:investigation}

\subsection{Common experimental setting}
Here, we describe the common setting in our experiments\footnote{Code is available at \href{https://github.com/mertan-a/pollination}{github.com/mertan-a/pollination}}. Unless otherwise noted, our experiments use the following setting: 

\textbf{Simulation environment} We use the Evogym~\cite{bhatia2021evolution} simulator to simulate voxel-based soft-bodied robots in 2D. Within this environment, robots are structured using a grid-like arrangement of voxels, comprising four distinct material types: passive soft materials, passive rigid materials, and active materials capable of actuating horizontally or vertically. In this work, we consider a 10-by-10 bounding box for robots. Similar to~\cite{bhatia2021evolution}, we assume that each voxel is connected to all its 4 adjacent neighbors to simplify the design space.

\textbf{Morphology representation and mutation} Robot morphologies are represented by a CPPN~\cite{stanley_compositional_2007} with sine, absolute value, square, and square root activation functions. Each CPPN generates five outputs, where four correspond to different material types and one signifies the absence of a voxel. The creation and material characteristics of a voxel are governed by the output node with the highest activation level. In the case of multiple disconnected patches, the biggest contiguous patch is kept and the others are discarded. CPPNs undergo mutation through the addition or removal of nodes or links, change of a node's activation function, or alteration of a link's weight. 

\begin{figure*}
    \centering
    \includegraphics[width=0.33\linewidth]{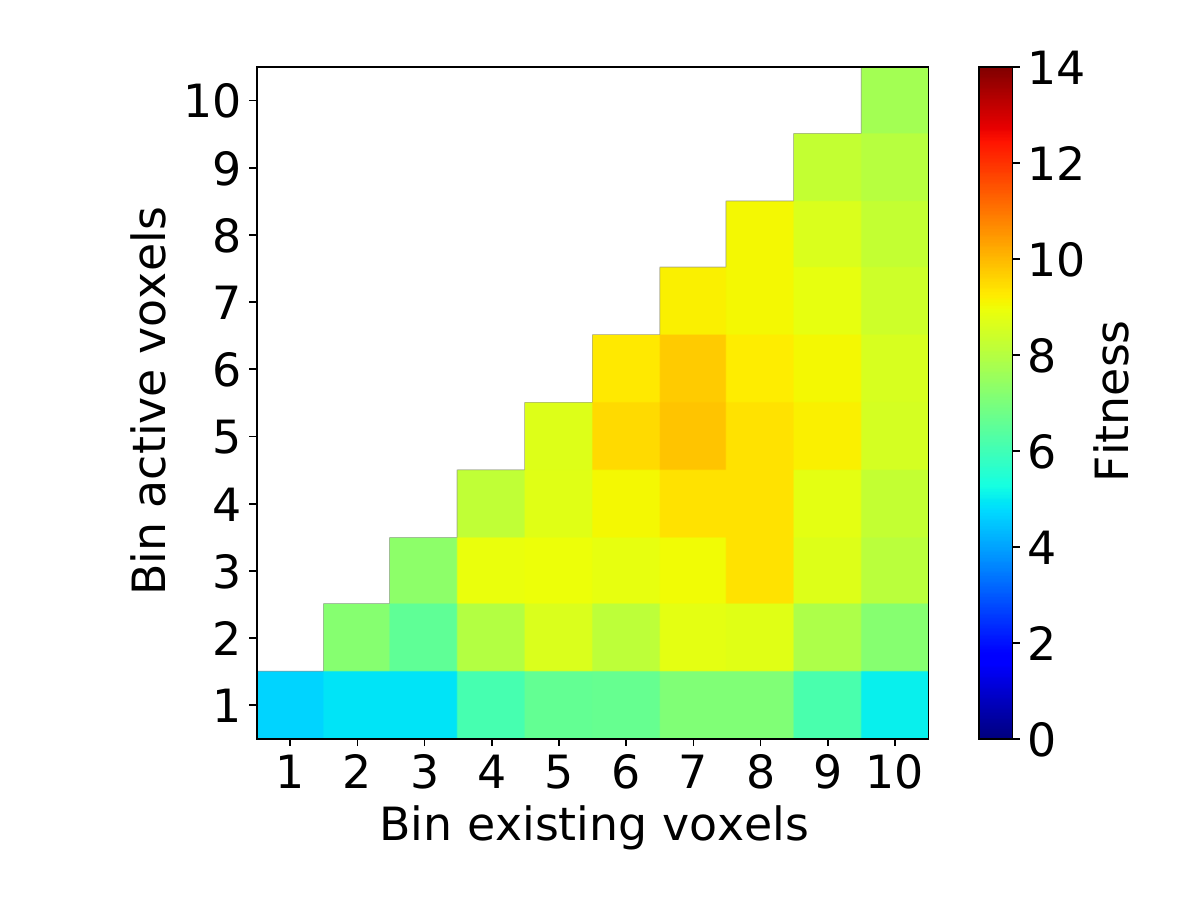}%
    \includegraphics[width=0.33\linewidth]{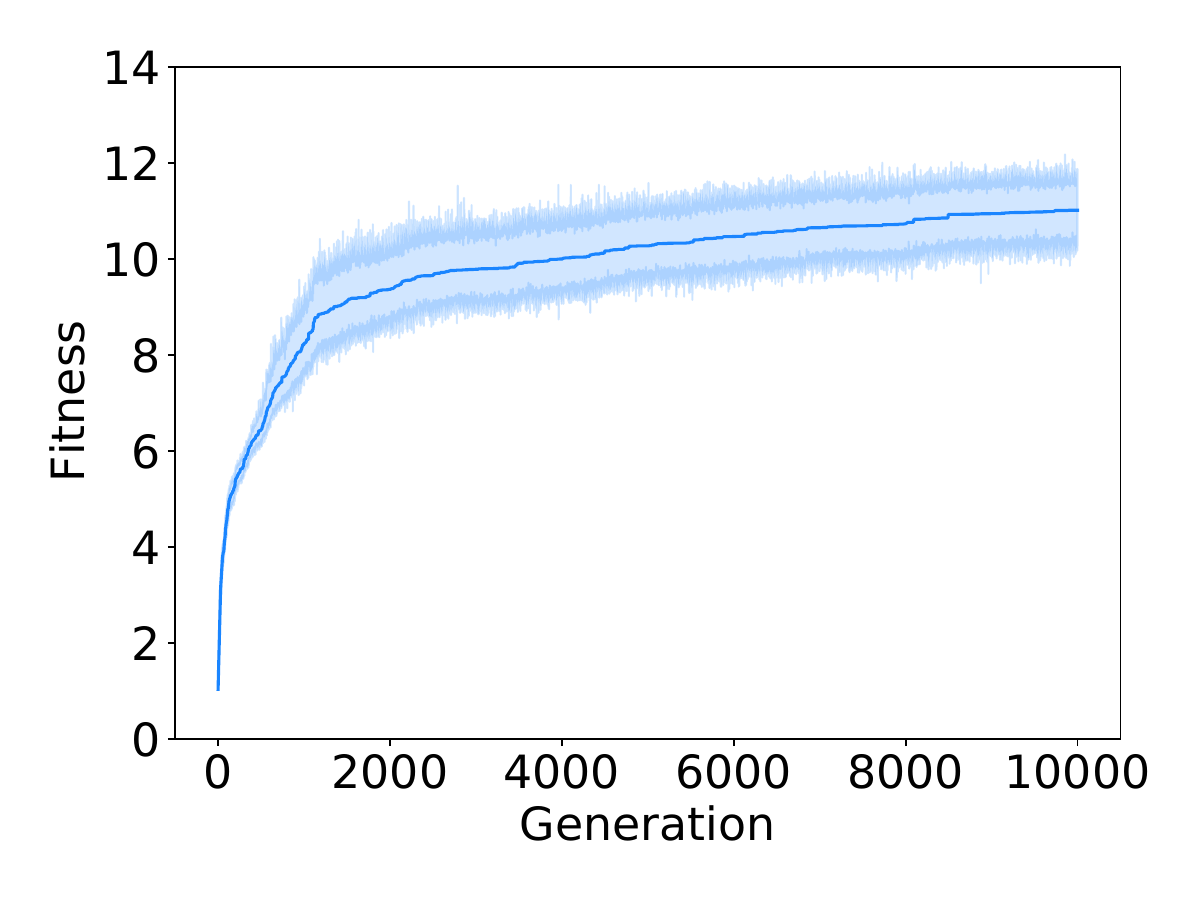}%
    \includegraphics[width=0.33\linewidth]{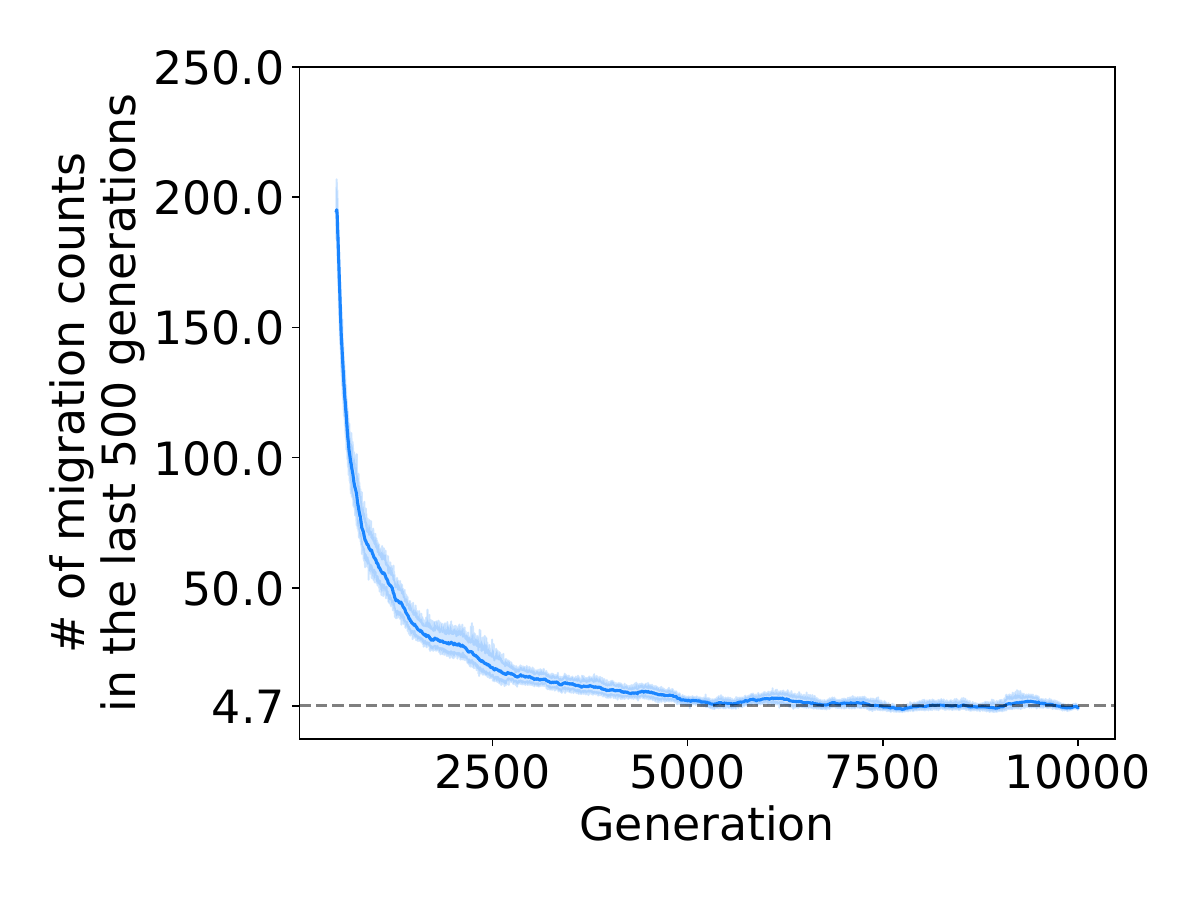}
    \vspace{-1em}
    \caption{Applying MAP-Elites algorithm directly, \textit{Standard} treatment, results in seemingly reasonable performance. Fitness values for each niche at the end of the run averaged over 20 repetitions (left) and the fitness of the best individual during evolutionary time (middle) shows reasonable performance. Inspecting the number of migrations, however, shows a trend where the number of migrations quickly decreases.  This observation raises concerns given that the strength of the MAP-Elites algorithm comes from its ability to create stepping stones. The solid lines show the averages over 20 repetitions, and the shaded regions show a 95\% confidence interval.
    }
    \label{fig:standard-results}
    \Description{desc}
\end{figure*}

\textbf{Controller representation and mutation} Robot brains are represented by a fully connected feedforward neural network with a single hidden layer of size 32. This network observes the volume, speed, and material properties of each voxel, and outputs actions representing the deformation target for each active voxel. Some noise sampled from $N(0, 0.01)$ is applied to both observations and actions to make the knowledge distillation process (described in Sect.~\ref{sect:pollination}) robust. In order to ensure the controller's compatibility across all potential morphologies within a 10-by-10 design space, the controller is designed to work with the biggest robot possible. For smaller robots, we fill in the observation for missing voxels with zeros and discard the outputs that do not correspond to an active voxel. The controllers are mutated by adding random noise sampled from $N(0, 0.1)$ to the weight vectors.

\textbf{Fitness function} We use the Walker-v0 environment, where the robot's task is to move toward the right side of the screen. 
We use a modified version of the fitness function described in the benchmark~\cite{bhatia2021evolution}, which rewards robots for moving towards right side of the screen at every time step and rewards them for reaching the goal position. We modify this fitness function by adding a small negative penalty at every time step to encourage robots to reach the goal position as fast as possible. We also add a positive constant reward to offset the penalty, ensuring that the resulting fitness values are solely positive for simplified analysis, as done similarly in~\cite{mertan_modular_2023}.
Individuals are evaluated five times under varying noise and the average fitness value of five evaluations is used as the final fitness score. 

\textbf{Evolutionary algorithm} We experiment with the MAP-Elites algorithm~\cite{mouret_illuminating_2015}. Similarly to~\cite{mouret_illuminating_2015,nordmoen2021map}, the archive comprises two dimensions, each containing 10 bins -- the number of total voxels in the robot and the number of active/muscle voxels in the robot. All experimented evolutionary algorithms are run for 10,000 generations. In each generation, 20 individuals are selected uniformly at random from the existing archive to be parents. Following~\cite{cheney_scalable_2018}, offspring are generated by mutating either the parent's morphology or the brain, each with equal likelihood. Additional details of each algorithm are described in the following sections. For comparing algorithms, we use the Wilcoxon rank sum test~\cite{wilcoxon1964some}. All comparisons are made with the $p = .05$ threshold.

\subsection{Applying MAP-Elites}

We apply the MAP-Elites algorithm~\cite{mouret_illuminating_2015} to the problem of brain-body co-optimization, similar to~\cite{nordmoen2021map}. This treatment is referred to as \textit{Standard}. Fig.~\ref{fig:standard-results} illustrates the results of the \textit{Standard} treatment. Fig.~\ref{fig:standard-results}, left, illustrates the fitness of each niche at the end of evolution, averaged over 20 independent repetitions. Compared to other work reporting results on the same simulator, task, and fitness function~\cite{mertan_modular_2023}, it seems that we are finding reasonably performing solutions (average run champion performance around 10). The fitness over time plot, illustrated in Fig.~\ref{fig:standard-results}, middle, also shows a very typical trend of fast improvement early on, followed by incremental improvements for the rest of the evolution, eventually converging on a solution (no fitness improvement larger than 0.05 in the last 2007.65 generations on average). On the other hand, Fig.~\ref{fig:standard-results}, right, shows a potentially worrying trend: the number of migrations (\emph{i.e.} solutions that move from one niche to another) decreases rapidly, averaging over 4.7 migrations per 500 generations for the majority of the evolution. The observation is concerning, as a strength of the MAP-Elites algorithm lies in its ability to create stepping stones, and the lack of migrations shown to hinder the performance of MAP-Elites~\cite{dean2023many}. And in this instance, we notice only a limited number of migrations occurring.

\begin{figure}
    \centering
    \includegraphics[width=0.5\linewidth]{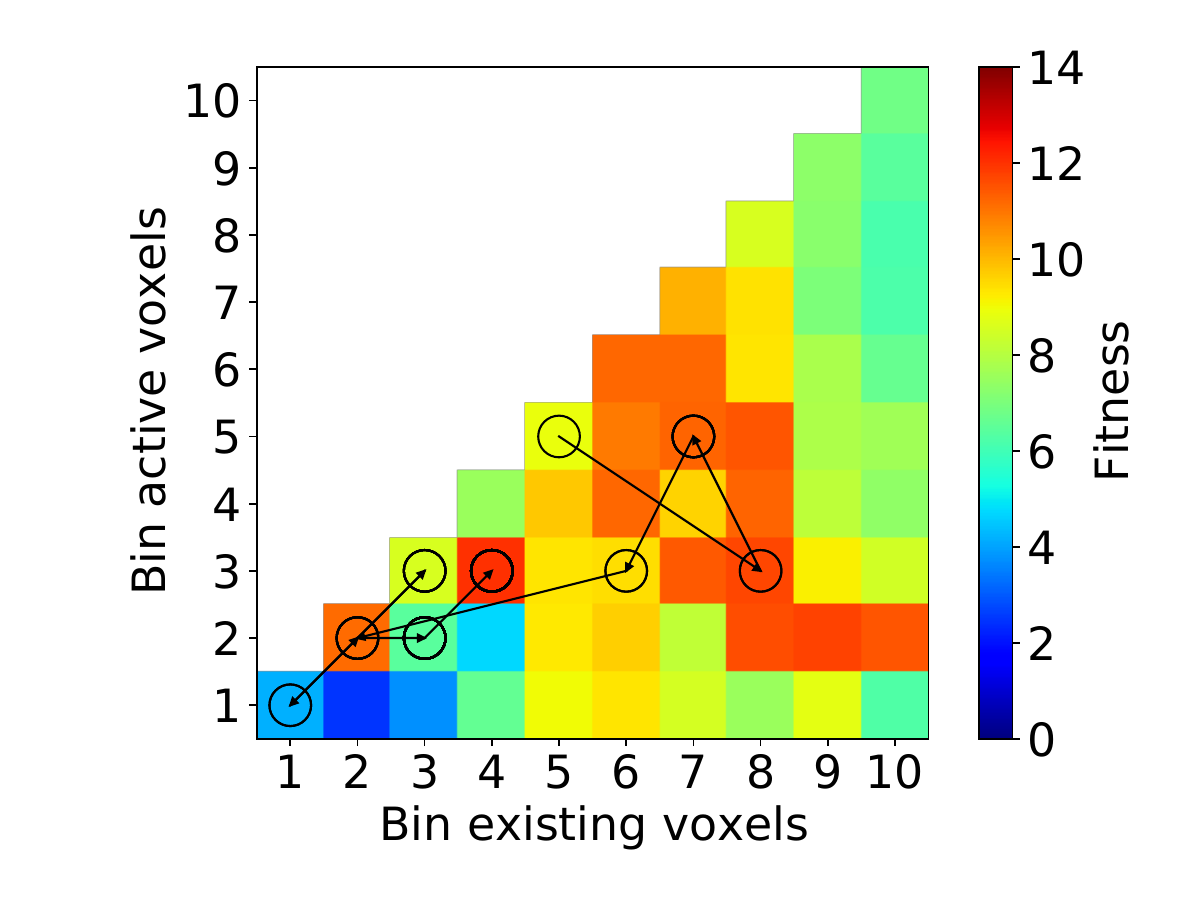}%
    \includegraphics[width=0.5\linewidth]{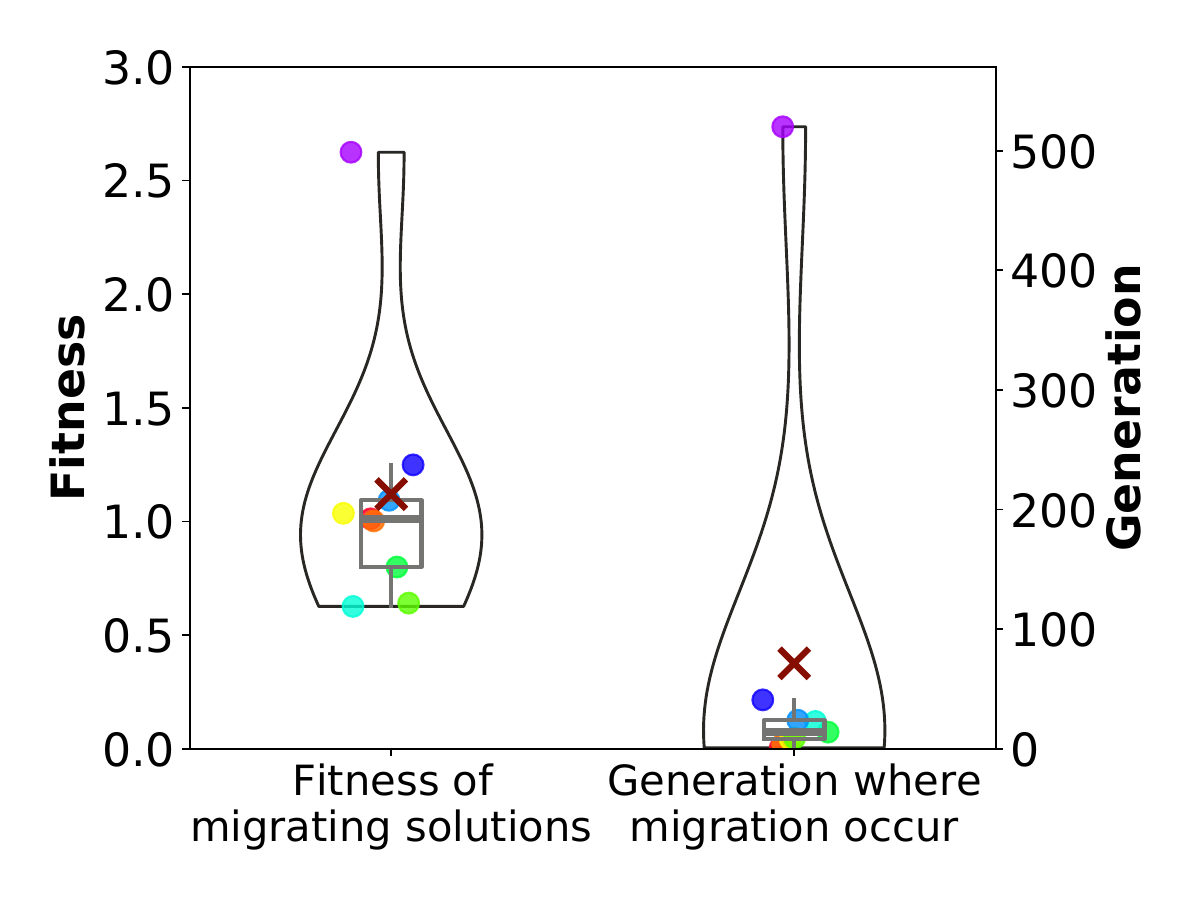}
    \caption{Archive of a selected run and the lineage of the run champion plotted on the niche space, where arrows show migration events (left). Fitness and generation of each migrating solution (right). While it seems like MAP-Elites is creating many stepping stones, these events happen at a low fitness regime and very early on in the evolution. Please note the scale on the y-axis.
    }
    \label{fig:standard-results-details}
    \Description{desc}
\end{figure}

\begin{figure*}
    \centering
    \includegraphics[width=0.33\linewidth]{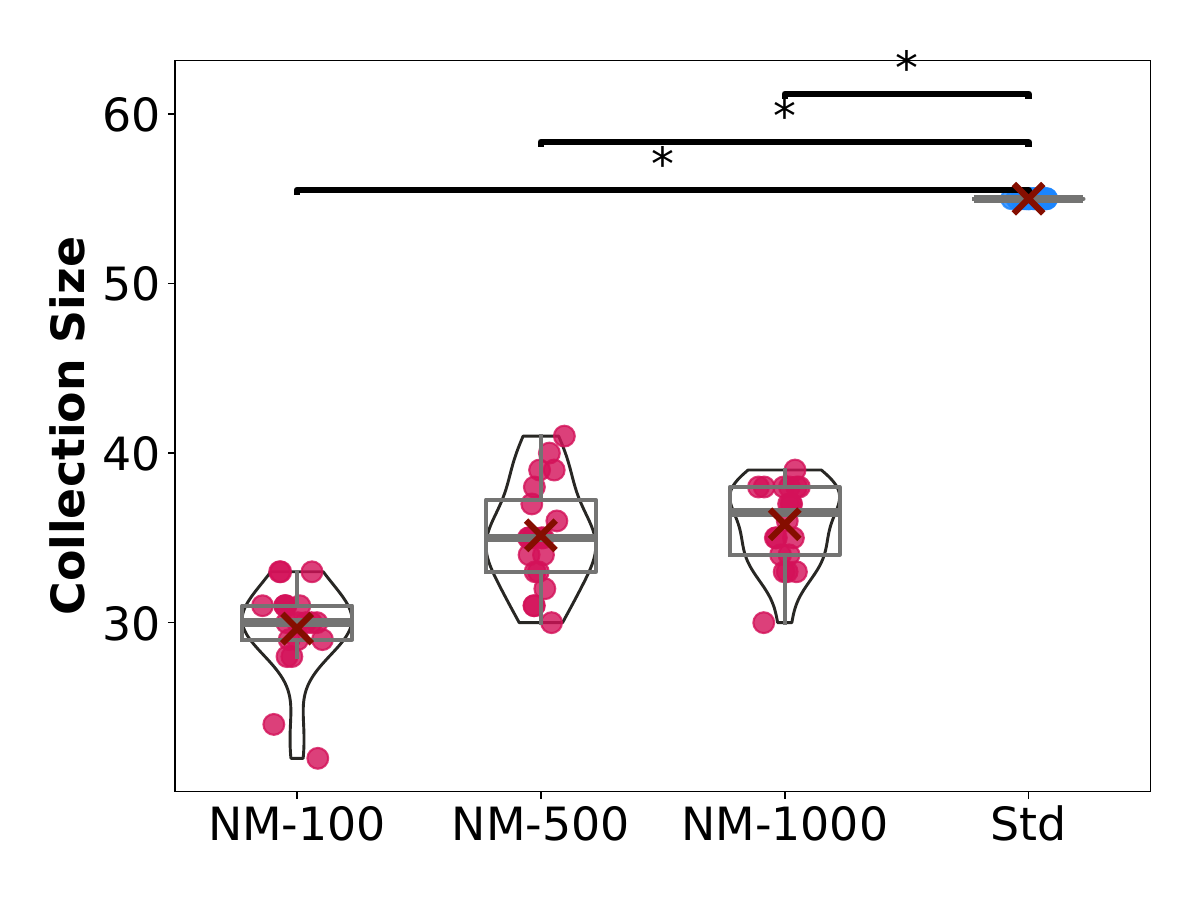}%
    \includegraphics[width=0.33\linewidth]{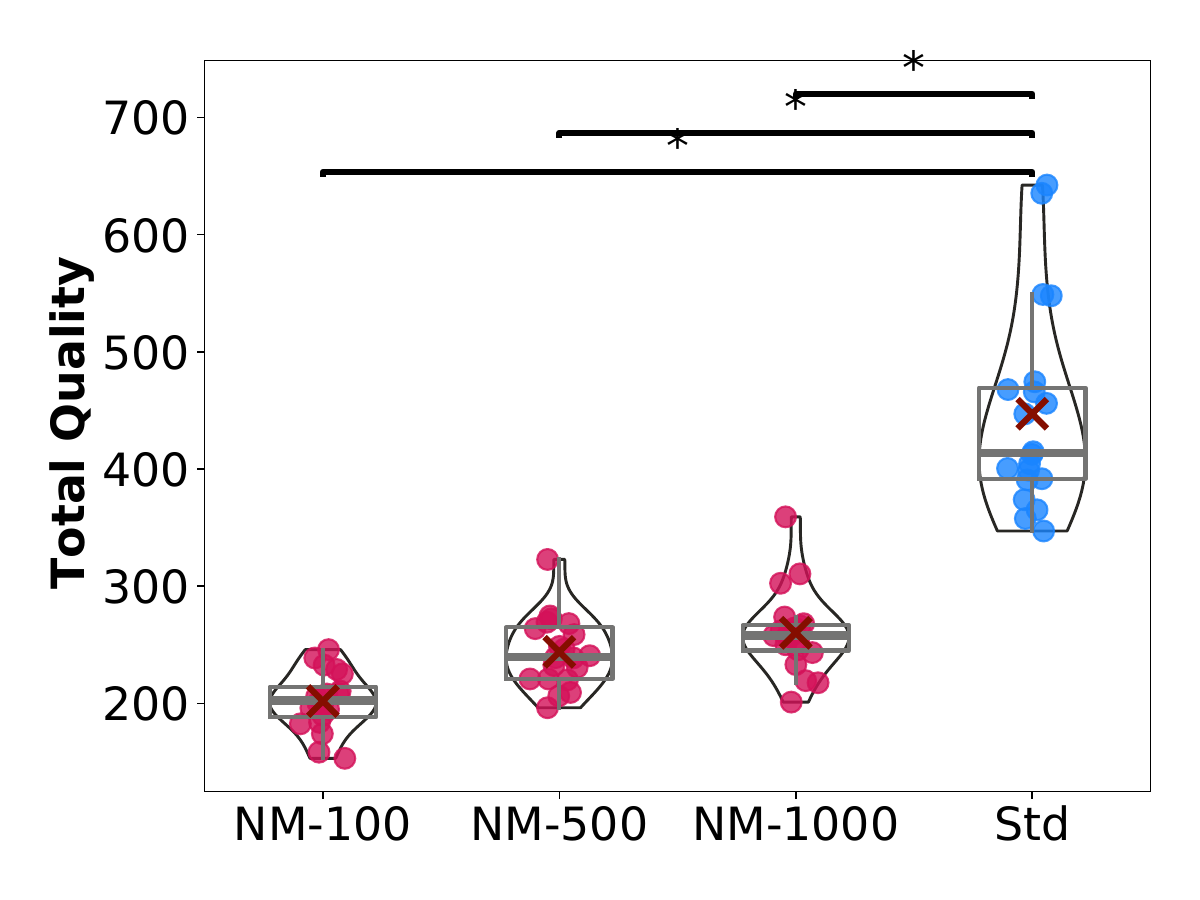}%
    \includegraphics[width=0.33\linewidth]{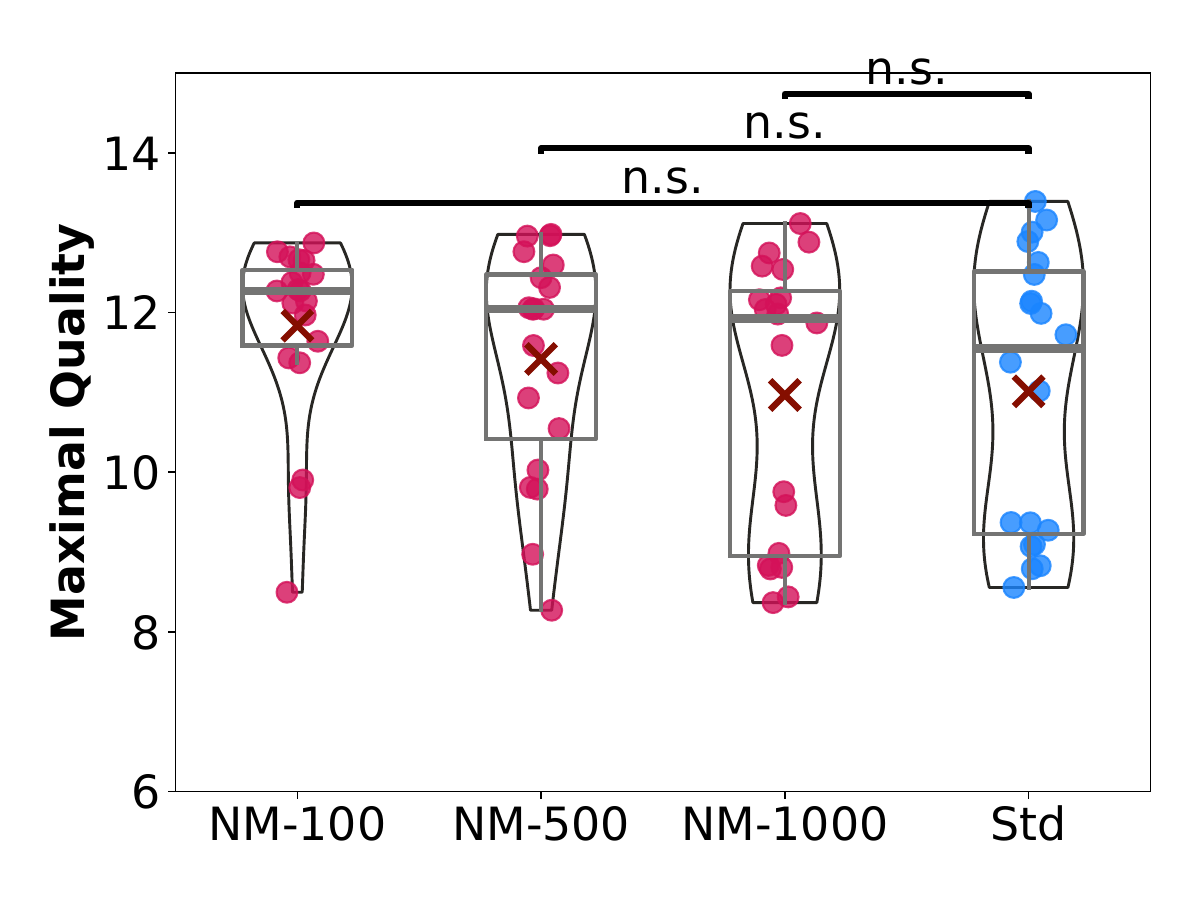}
    \vspace{-1em}
    \caption{Comparison of quality metrics for the \textit{Standard} and \textit{No Migration} treatments. \textit{Standard} treatment consistently fills up all of the available niches while the \textit{No Migration} treatment loses the morphological diversity as there is no explicit mechanism to protect it, which results in better collection size and total quality for the \textit{Standard} treatment. However, both treatments perform comparably in maximal quality, prompting us to question whether we could have found better solutions with more migrations. Horizontal lines shows the results of statistical tests. *: $P<0.05$, n.s.: $P>=0.05$}
    \label{fig:compare-std-control}
    \Description{desc}
\end{figure*}

There is, of course, no a priori number of migrations that needs to occur, or a particular trend that the number of migrations over evolutionary time should follow. In fact, it makes sense to see a decrease in the number of migrations over evolutionary time since migrations become harder as evolution optimizes solutions more and more. However, seeing only a handful of migrations per 500 generations prompted us to investigate this phenomenon further. Fig.~\ref{fig:standard-results-details} provides a detailed analysis of a selected run. Fig.~\ref{fig:standard-results-details}, left, shows the migrations that happen in the lineage of the run champion, plotted over the niche space, where each node represents an ancestor and the arrows represent migrations. MAP-Elites is capable of creating many stepping stones on the road to finding the run champion, which is in agreement with the findings of~\cite{nordmoen2021map}. However, as shown in Fig.~\ref{fig:standard-results-details}, right, these migrations occur at a low fitness regime (about 20\% of the run champion's fitness and lower; average of 9.4\%) and in early generations (none in the last 90\% of the time it takes to find the run champion, on average in the first 0.9\%). This result leads us to further scrutinize the importance of the rare migrations that happen at later stages of evolution.

\subsection{Control with no migration}

To investigate the usefulness of migrations that we observe, we devise a control named \textit{No Migration} where we turn the MAP-Elites algorithm into a parallel hillclimber after $N$ number of generations. We run the MAP-Elites algorithm normally until $N$\textsuperscript{th} generation. After $N$\textsuperscript{th} generation, each solution in the archive spawns a hillclimbing process. To keep the comparison fair in terms of the number of fitness evaluations, we iterate 20 parallel hillclimber processes at each generation.

Fig.~\ref{fig:compare-std-control} compares the \textit{Standard} treatment against \textit{No Migrations} after 100, 500, or 1000 generations on the quality metrics described in~\cite{cully2017quality}. The \textit{Standard} treatment is capable of consistently filling all niches (i.e. collection size $=$ map size). On the other hand, the \textit{No Migration} treatments have not yet filled all available niches before migrations are eliminated. In addition, they lose diversity in the hillclimbing process in the absence of an explicit niching mechanism to maintain morphological diversity. This results in significantly better collection sizes (number of niches filled) and total quality (sum of fitness across all filled niches) metrics for the \textit{Standard} treatment. However, both treatments perform comparably in terms of the maximal quality (most fit individual found), with no significant difference in the quality of the best robot found from the \textit{Standard} treatment with full migrations compared to the treatments with \textit{No Migrations} after as few as 100 out of the 10,000 total generations. The \textit{Standard} treatment is unable to utilize later migrations to find better run champions and cannot effectively outperform parallel hillclimbing.

Given the previous work that has explored the brain-body co-optimization problem and its challenges~\cite{cheney_difficulty_2016,mertan2024investigating}, namely fragile co-adaptation in which brains become over-specialized for particular morphologies and do not transfer well to other morphologies, it is not surprising to see a limited number of migrations. Moreover, it seems like in the traditional application of MAP-Elites for brain-body co-optimization, migrations that happen later on in evolution do not help with the quality of the run champions, negating the algorithm's strength of avoiding local optima through the creation of diverse stepping stones. 
Thus 
we hypothesize that promoting migrations will enable MAP-Elites to work better and to discover better solutions. So, the question is, how can we encourage more migrations in brain-body co-optimization?

\section{A Potential Improvement}\label{sect:pollination} 

\begin{table}
    \centering
    \begin{tabularx}{\linewidth}{lX}
        \hline
        \textbf{Hyperparameter} & \textbf{Description: Implementation} \\
        \hline
        \rule{0pt}{3ex}    
        \hypertarget{PF}{Pollination Frequency} & The number of generations to perform the pollination process: 500 \\
        \rule{0pt}{3ex}    
        \hypertarget{TS}{Teacher Selection} & The strategy to choose solutions from the archive for knowledge distillation: one randomly selected solution and its 4 Moore neighbors in niche space \\
        \rule{0pt}{3ex}    
        \hypertarget{DP}{Distillation Parameters} & It includes the training algorithm, optimizer, learning rate, and number of epochs to train: same as~\cite{mertan2024towards} \\
        \rule{0pt}{3ex}    
        \hypertarget{PS}{Pollinated Selection} & The strategy to choose solutions from the archive to replace their controllers with the distilled one: teacher solutions are pollinated if the distilled controller performs comparably to the teacher controller \\
        \rule{0pt}{3ex}    
        \hypertarget{PE}{Pollinated Evaluation} & The strategy to determine the fitness of the pollinated individuals: re-evaluate with the distilled controller \\
        \hline
    \end{tabularx}
    \caption{Hyperparameters of the pollination process. Apart from distillation parameters, all hyperparameters are empirically determined based on limited preliminary results.}
    \label{tab:hyperparameters}
\end{table}

To alleviate fragile co-adaptation and enable migrations, we investigate how we can empower controllers so that they can transfer better to other morphologies in a zero-shot manner -- i.e. how a controller can be robust to morphological changes instead of suffering from fragile co-adaptation. In our previous work~\cite{mertan2024towards}, we have demonstrated a potential solution to the problem of finding more robust controllers.  We show that knowledge distillation~\cite{buciluǎ2006model,hinton2015distilling,rusu2015policy,parisotto2015actor} from a set of expert teacher controllers (in which each teacher is optimized for a different morphology)
is capable of creating controllers as effective as the teacher controllers in controlling the teacher morphologies, and these controllers exhibit proficiency in generalizing to new morphologies in a zero-shot manner. 



Here, we employ this method to create empowered controllers. We refer to this process as \textit{Pollination}, since it involves the exchange of information among individuals to produce offspring with greater fitness. 
\textit{Pollination} involves some choices and hyperparameters, which are illustrated in Table~\ref{tab:hyperparameters}. We have investigated only a handful of options to demonstrate the efficiency of the proposed method as a proof-of-concept and left a detailed analysis of potential options and their analysis for each of the hyperparameters for future work.

Specifically, in \textit{Pollination} treatment, every \hyperlink{PF}{\textit{Pollination Frequency}} (see Table~\ref{tab:hyperparameters}) number of generations during evolution, we sample teacher morphology-controller pairs (\hyperlink{TS}{\textit{Teacher Selection}}) from the archive and collect observation-action pairs in a shared dataset by running each teacher. Supervised training is then used to optimize a single controller to mimic teacher policies (\hyperlink{DP}{\textit{Distillation Parameters}}). Lastly, we replace some of the controllers in the MAP-Elites archive with these distilled empowered controllers (\hyperlink{PS}{\textit{Pollinated Selection}} and \hyperlink{PE}{\textit{Pollinated Evaluation}}).

\begin{figure}
    \centering
    \includegraphics[width=0.8\linewidth]{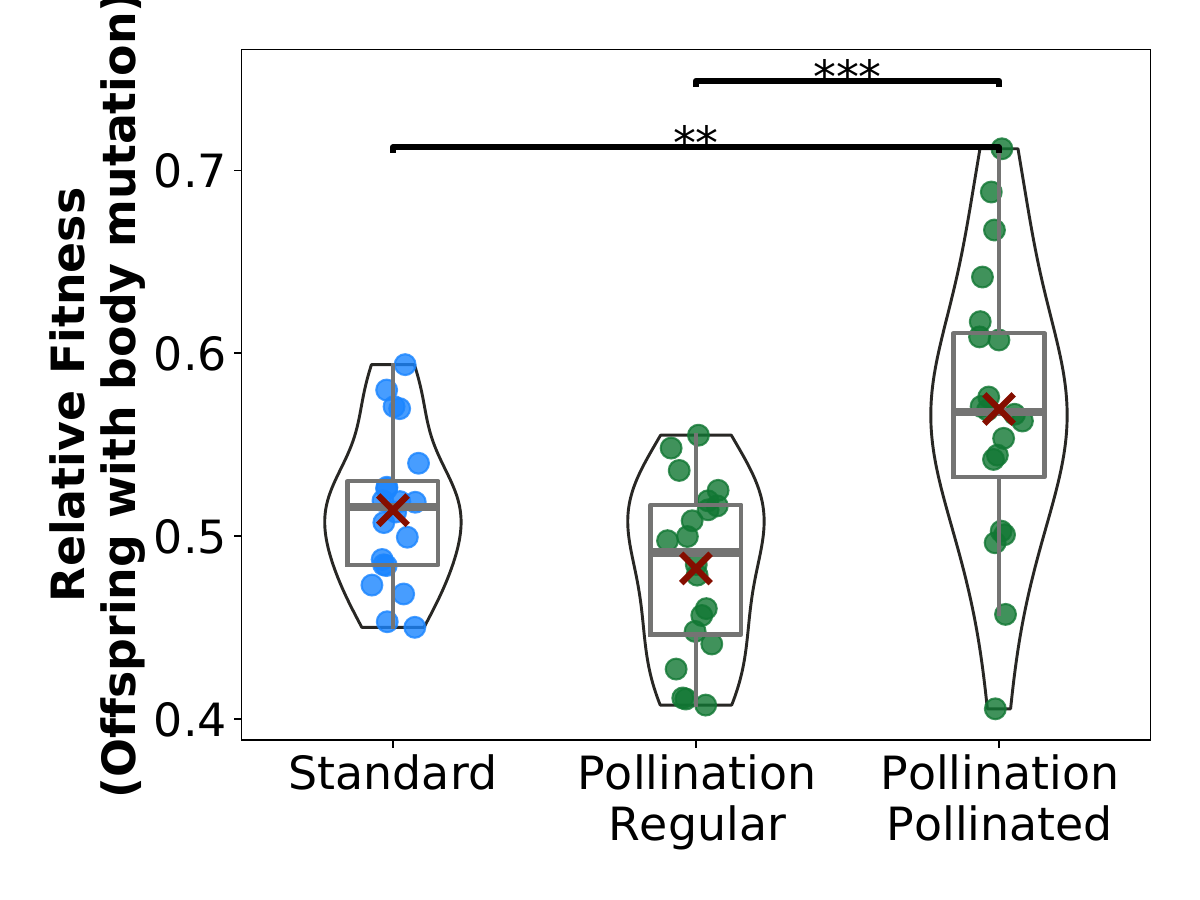}
    \caption{Distribution of the average relative fitness of the offspring created through body mutations relative to their parents at each repetition of the \textit{Standard} and \textit{Pollination} treatments. Pollination Regular covers the offspring with no pollinated ancestor -- none of their ancestors have had their controllers replaced with a distilled controller. Pollination Pollinated covers the offspring with at least one pollinated ancestor in their lineage. The distributions of the average relative fitness for the \textit{Standard} treatment and the regular offspring in the \textit{Pollination} treatment are statistically indistinguishable at level $p>0.05$. On the other hand, offspring with pollinated ancestors are more robust to body mutations compared to other groups, confirming that distilled controllers transfer better to other morphologies and can potentially encourage more migrations. }
    \label{fig:body-mutation-fitness-change-detailed}
    \Description{desc}
\end{figure}

We start analyzing the \textit{Pollination} treatment by investigating whether it achieves what we set out to do: encouraging more migrations. Since the niche space is designed based on morphological features, only offspring created through body mutations can migrate from one niche to another. In Fig.~\ref{fig:body-mutation-fitness-change-detailed}, we analyze the fitness of the offspring produced by body mutations relative to their parents. We compare the \textit{Standard} treatment, offspring with no pollinated ancestor from the \textit{Pollination} treatment, and offspring with at least one pollinated ancestor from the \textit{Pollination} treatment. The results illustrate that, while the offspring in the \textit{Standard} treatment and the offspring without pollinated ancestor in the \textit{Pollination} treatment have statistically indistinguishable relative fitness on average, the offspring with at least one pollinated ancestor significantly outperforms both groups. These results show that the proposed method can create solutions that are more robust to morphological mutations.

\begin{figure*}
    \centering
    \includegraphics[width=0.33\linewidth]{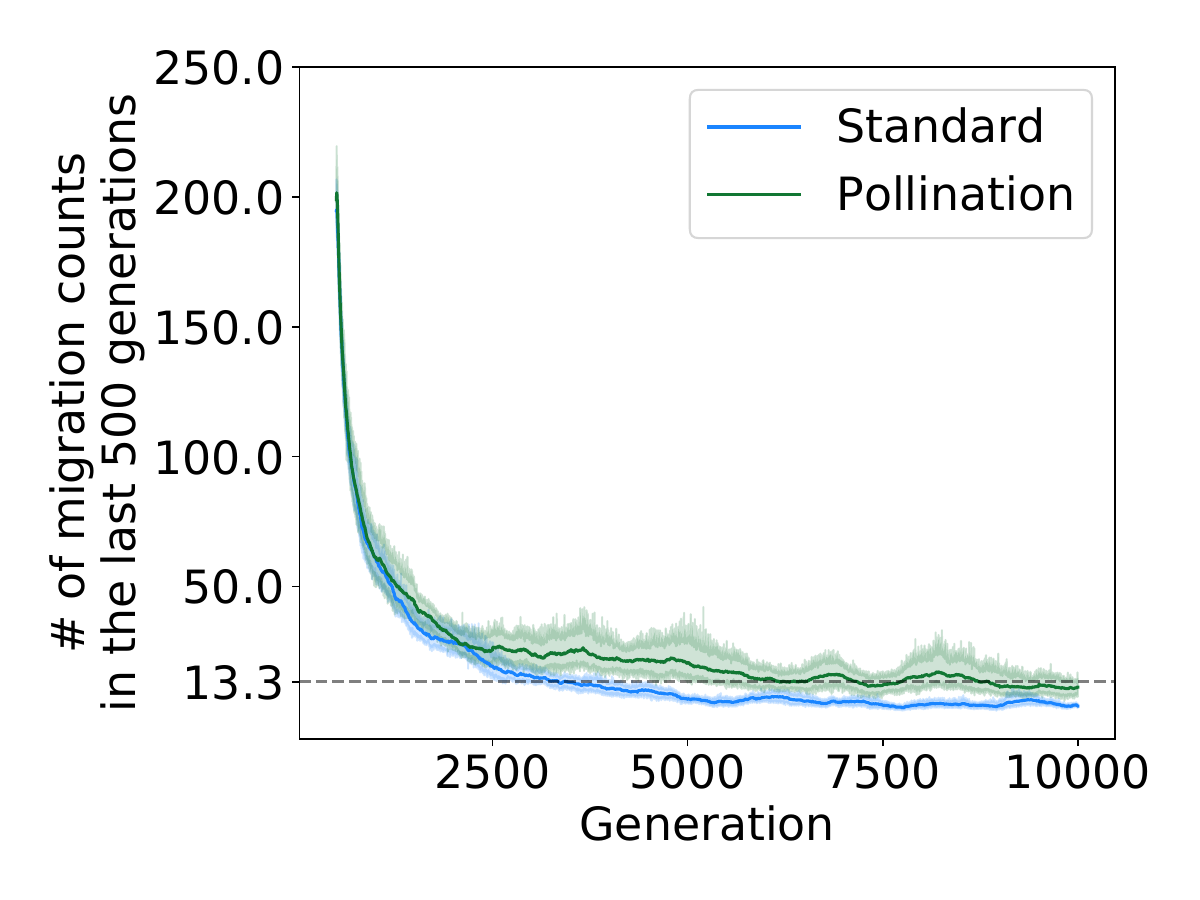}%
    \includegraphics[width=0.33\linewidth]{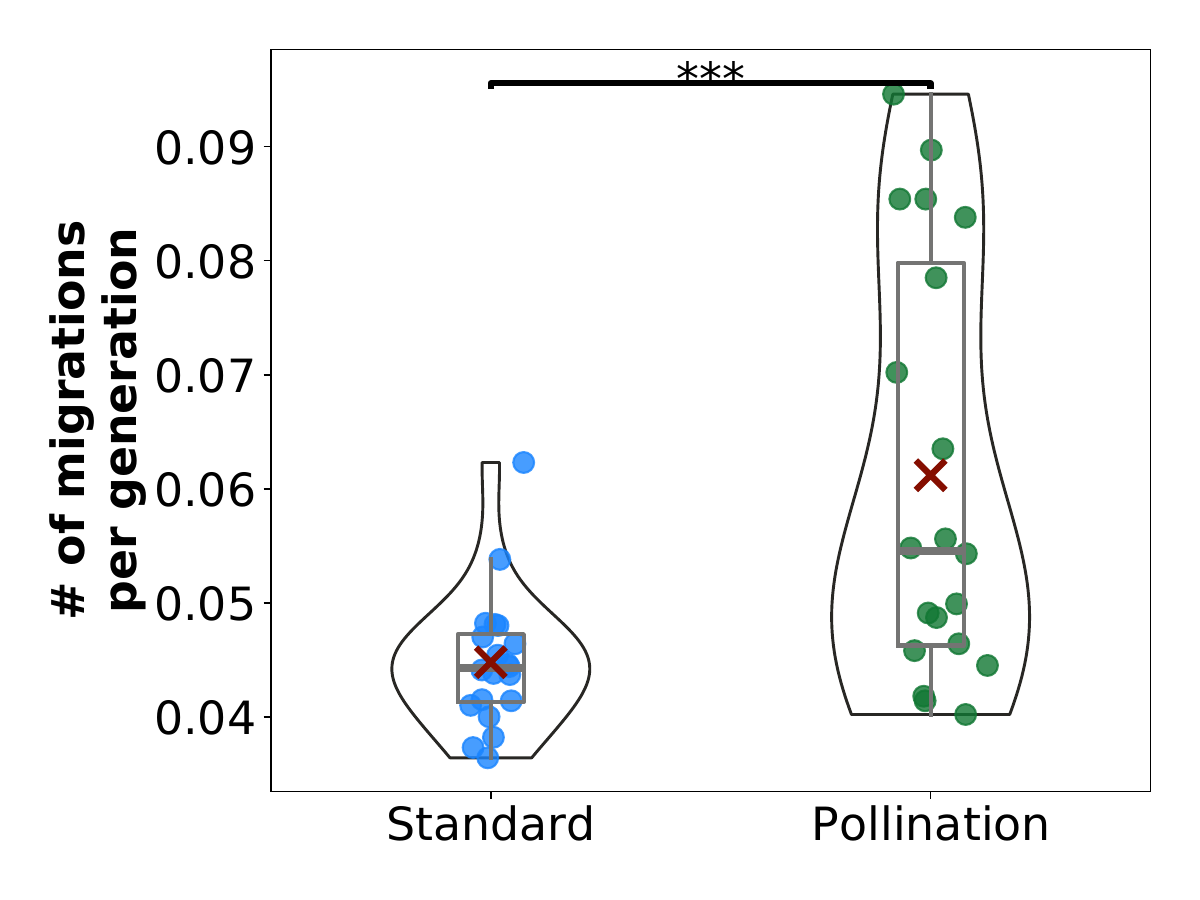}%
    \includegraphics[width=0.33\linewidth]{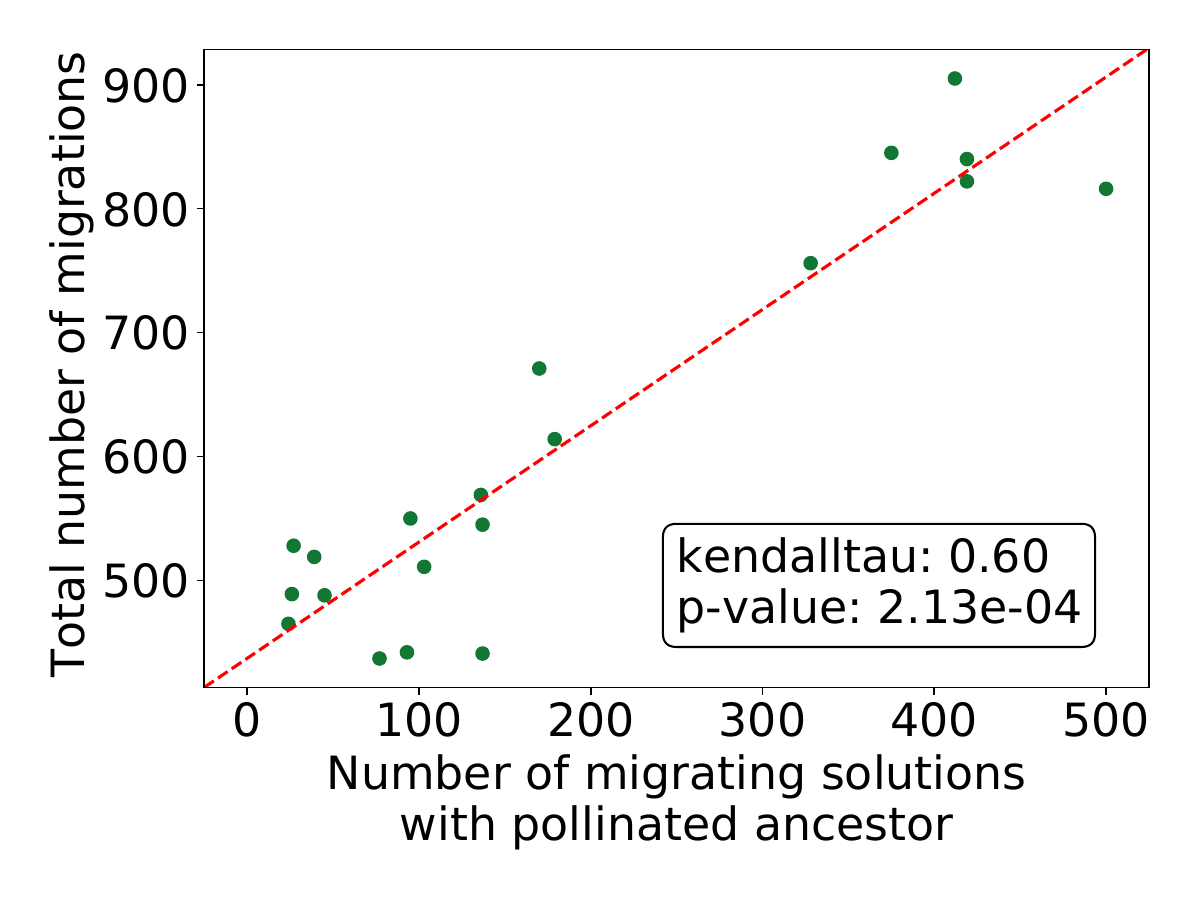}
    \vspace{-1em}
    \caption{We investigate the migrations occurred in the \textit{Standard} and \textit{Pollination} treatments. Migrations over evolutionary time (left) and distributions of the average number of migrations per generation for each repetition (middle) are plotted. While both treatments show a similar trend over evolutionary time, \textit{Pollination} treatment is capable of producing more migrations on average during whole evolution and on average in total. The total number of migrations is plotted against the number of migrations where the migrating solution has at least one pollinated ancestor, for each repetition of the \textit{Pollination} treatment (right). The strong correlation shows that the migration increase results from the proposed pollination procedure.}
    \label{fig:compare-std-pollination-migration}
    \Description{desc}
\end{figure*}

\begin{figure}
    \centering
    \includegraphics[width=0.5\linewidth]{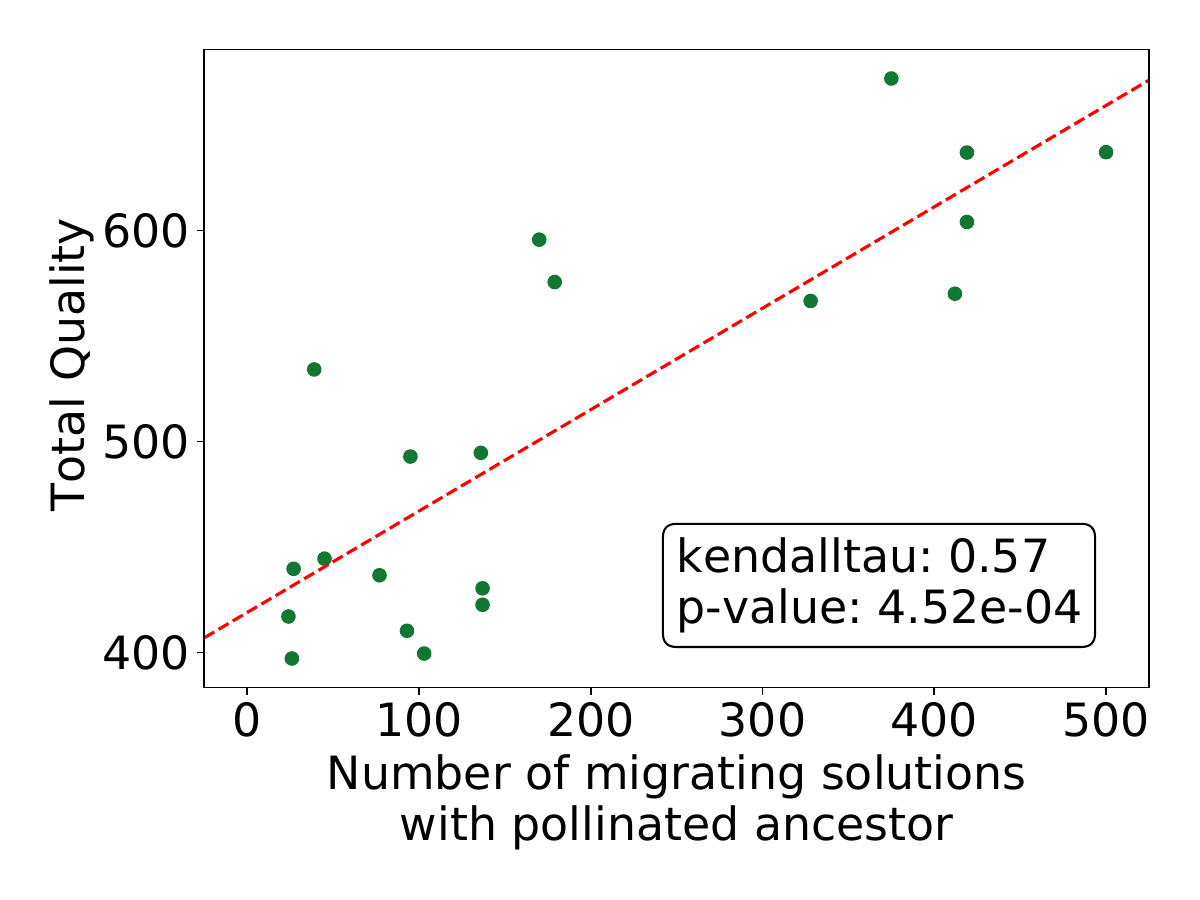}%
    \includegraphics[width=0.5\linewidth]{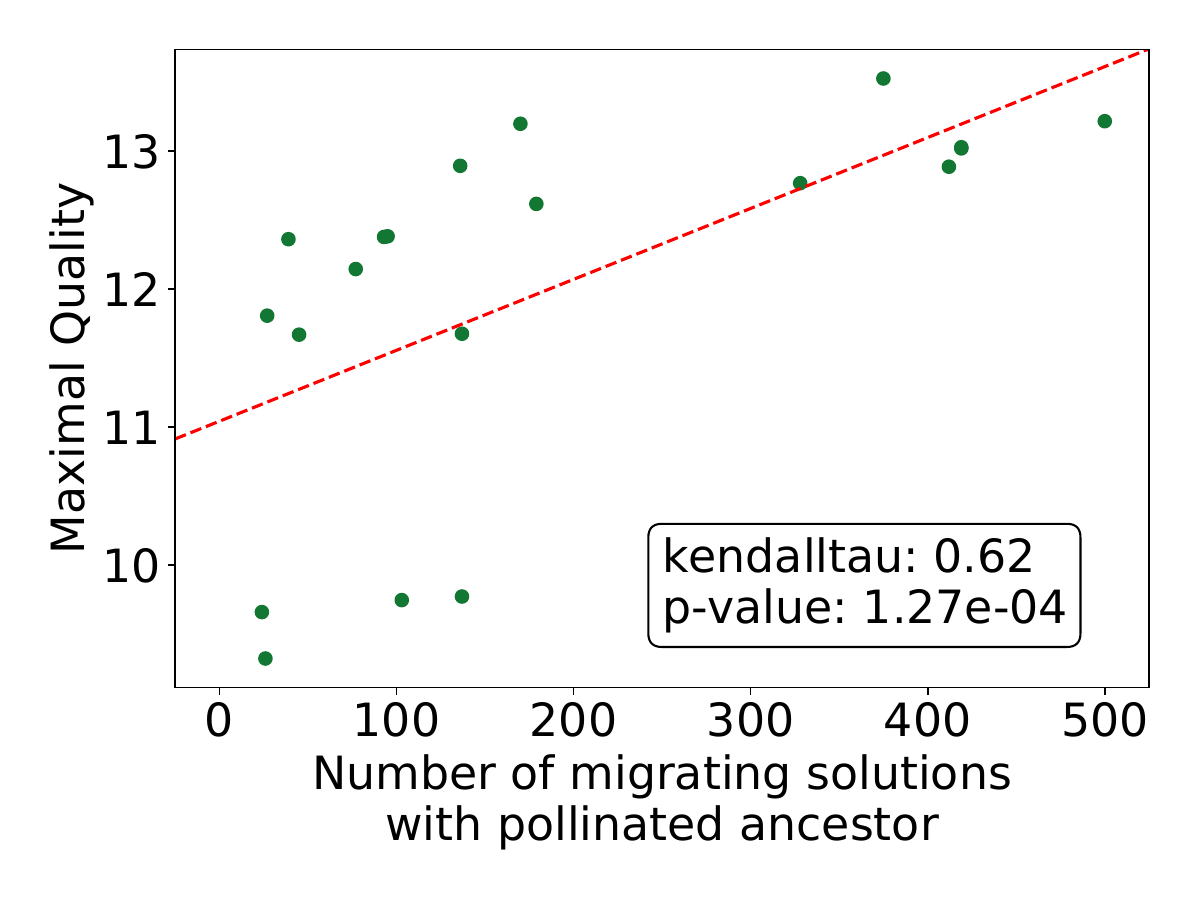}
    \caption{Total quality (left) and maximal quality (right) of each repetition of the \textit{Pollination} treatment, plotted against the number of migrations where the migrating solution has at least one pollinated ancestor. Both plots strongly correlate quality-diversity metrics and pollination-induced migrations, strengthening our hypothesis that the proposed pollination method will enable a better search.}
    \label{fig:pollination-performance-vs-migration}
    \Description{desc}
\end{figure}

\begin{figure*}
    \centering
    \includegraphics[width=0.33\linewidth]{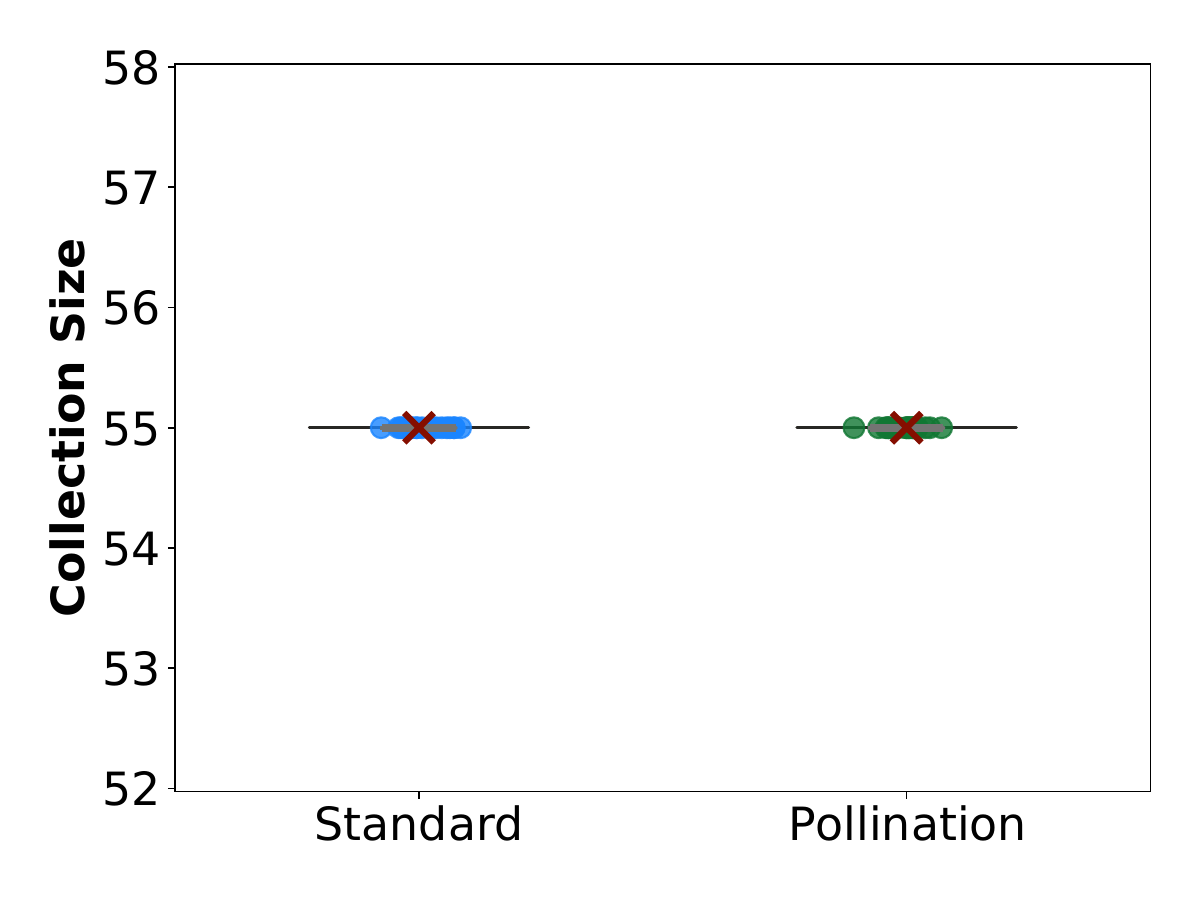}%
    \includegraphics[width=0.33\linewidth]{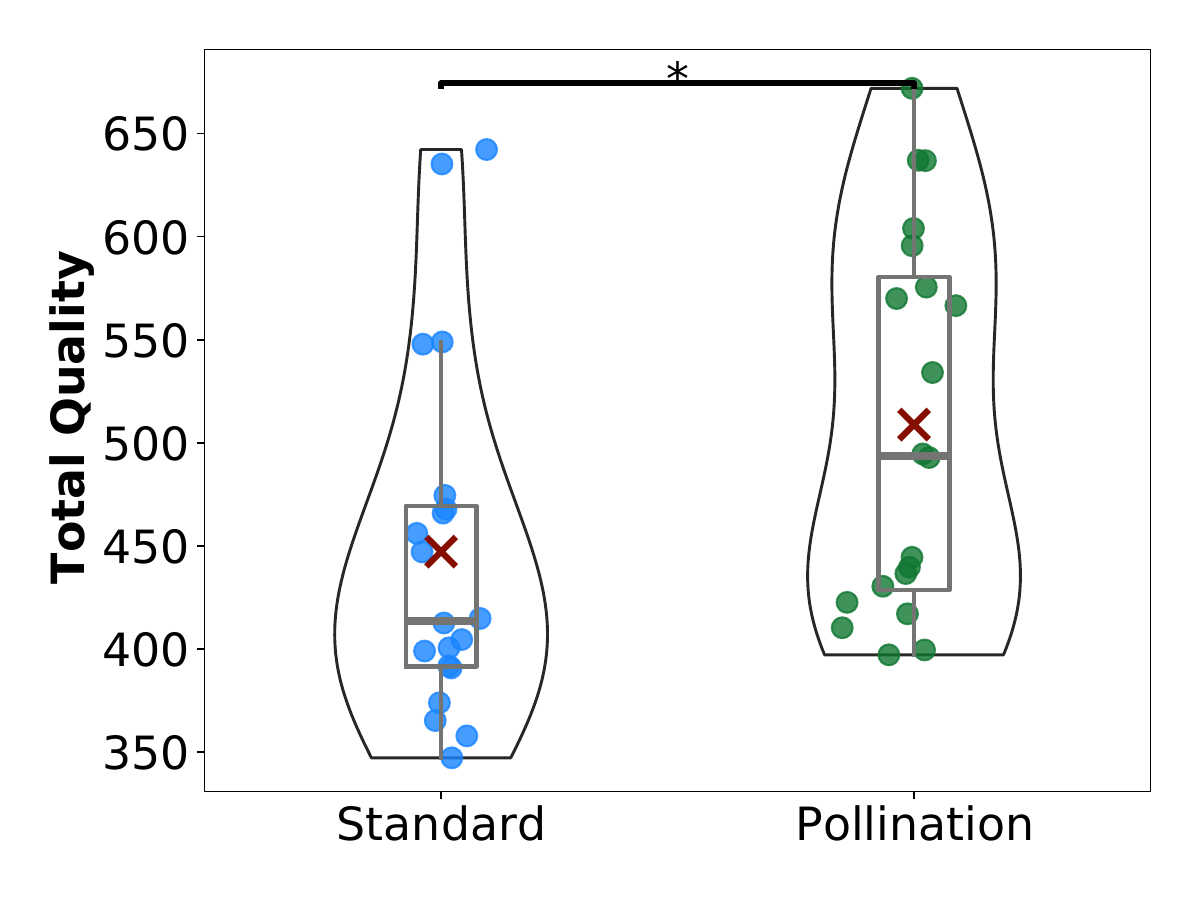}%
    \includegraphics[width=0.33\linewidth]{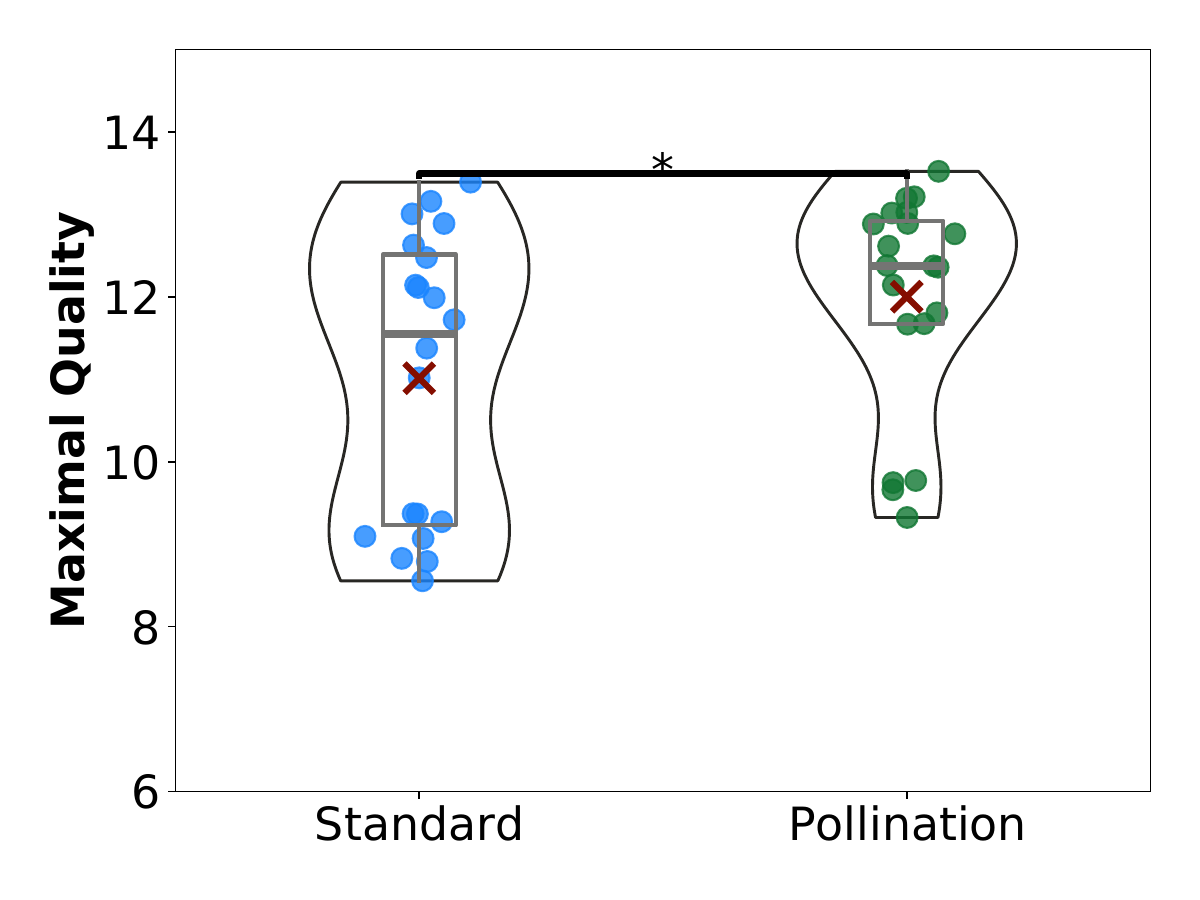}
    \vspace{-1em}
    \caption{Comparison of quality-diversity metrics for the \textit{Standard} and the proposed \textit{Pollination} treatments. Both treatments consistently fill up all of the available niches. \textit{Pollination} outperforms the \textit{Standard} treatment in both total quality and maximal quality.}
    \label{fig:compare-std-pollination}
    \Description{desc}
\end{figure*}

Next, Fig.~\ref{fig:compare-std-pollination-migration} plots migrations over evolutionary time (left) and in an aggregated way (middle). The results clearly show that more migrations happen in the \textit{Pollination} treatment compared to the baseline \textit{Standard} treatment. Although both treatments show the same trend of rapidly decreasing migrations over evolutionary time, we see that a better response to morphological mutations (provided by the pollination process, shown in Fig.~\ref{fig:body-mutation-fitness-change-detailed}) does translate to more migration on average throughout evolution. We also plot the relation between the total number of migrations and the number of migrations where the migrating solution has at least one pollinated ancestor (Fig.~\ref{fig:compare-std-pollination-migration}, right). The strong correlation between the two suggests that the increase in the number of migrations is indeed a result of the pollination process.

Having shown the relation between the proposed pollination method and the increased number of migrations, we focus on the second part of our hypothesis -- increased migrations induced by the pollination process will increase the success of the search algorithm. To this end, we examine how migrations induced by pollinated individuals affect the final results in the \textit{Pollination} treatment. Fig.~\ref{fig:pollination-performance-vs-migration} plots the percentage of migrating solutions with pollinated ancestors against the resulting total quality (Fig.~\ref{fig:pollination-performance-vs-migration}, left) and maximal quality (Fig.~\ref{fig:pollination-performance-vs-migration}, right) for each repetition. The results show a strong correlation between the presence of pollinated-induced migrations and quality-diversity metrics, strengthening our hypothesis that the proposed method will enable a better search.

Lastly, Fig.~\ref{fig:compare-std-pollination} compares the \textit{Standard} and \textit{Pollination} treatments on the quality-diversity metrics. Both treatments explore the niche space successfully, consistently discovering solutions for each available niche in all 20 runs (Fig.~\ref{fig:compare-std-pollination}, left). On the other hand, the \textit{Pollination} treatment is capable of finding better solutions across the niche space, indicated by higher total quality (Fig.~\ref{fig:compare-std-pollination}, middle). Moreover, we see better run champions in the \textit{Pollination} treatment (Fig.~\ref{fig:compare-std-pollination}, right). The distribution of run champions' fitnesses found in the \textit{Pollination} treatment demonstrates better mean, median, max, and min performances.

\begin{figure*}
    \centering
    \includegraphics[width=0.5\linewidth]{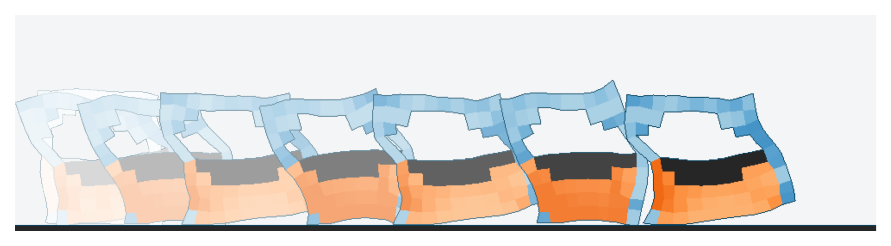}%
    \includegraphics[width=0.5\linewidth]{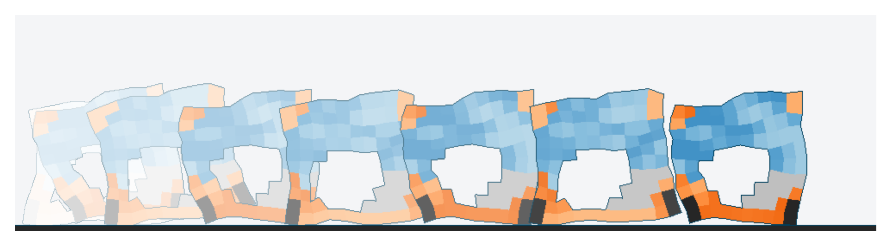} 
    \vspace{-0.6cm}
    \\
    \includegraphics[width=0.5\linewidth]{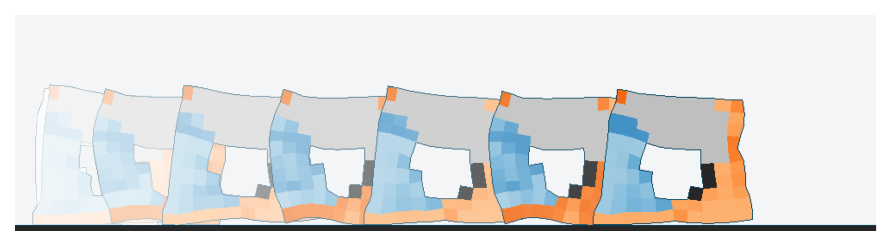}%
    \includegraphics[width=0.5\linewidth]{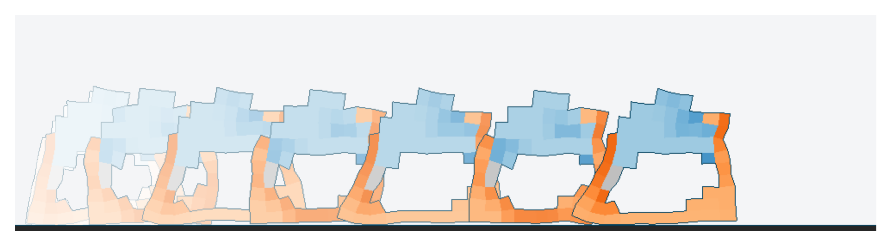} 
    \vspace{-0.6cm}
    \\
    \includegraphics[width=0.5\linewidth]{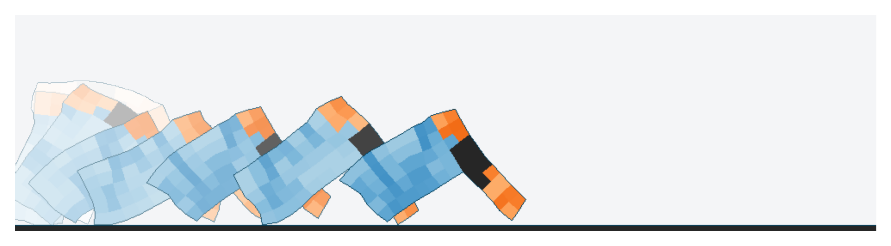}%
    \includegraphics[width=0.5\linewidth]{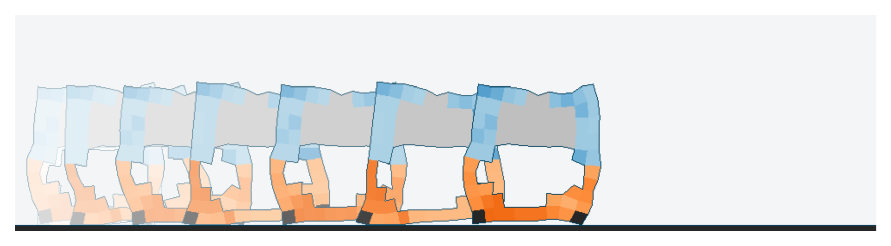} 
    \vspace{-0.6cm}
    \\
    \includegraphics[width=0.5\linewidth]{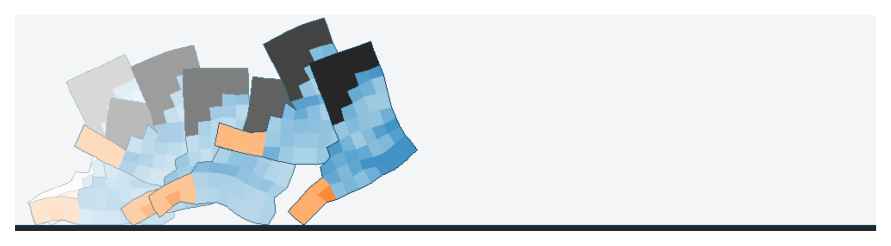}%
    \includegraphics[width=0.5\linewidth]{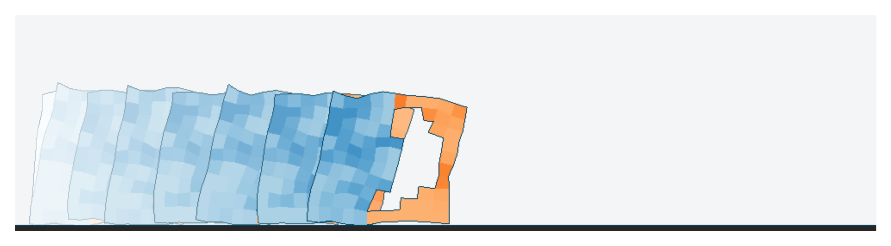}%
    \vspace{-1em}
    \caption{Behavior collages of run champions from the \textit{Standard} (left) and \textit{Pollination} (right) treatments. Images are ordered in descending order of their fitness from top to bottom where the top three images show the three best run champions found in all repetitions, while the last row shows the worst run champion from both treatments, out of 20 repetitions. High-performing run champions show similar designs of an empty square where robots move by arching their bottom. On the other hand, we see more diversity in the run champions with the \textit{Standard} treatment, while the \textit{Pollination} treatment more consistently finds the empty square design. Note that diversity in run champions coming from independent repetitions suggests a poor search as the algorithm fails to find and exploit the high-performing regions of the search space.}
    \label{fig:behavior-collages}
    \Description{desc}
\end{figure*}

Fig.~\ref{fig:behavior-collages} plots images from a lifetime of run champions from both \textit{Standard} and \textit{Pollination} treatments, to give an idea of the gaits and designs that are found by both treatments. The left column shows the run champions from the \textit{Standard} treatment and the right column shows the \textit{Pollination} treatment's run champions. The top three rows show the three best-performing run champions out of 20 independent runs, and the last row shows the worst-performing run champions. The prominent morphology design among the run champions appears to be an empty square (2 out of 20 in the \textit{Standard} treatment and 7 out of 20 in the \textit{Pollination} treatment). This morphology design also is the best-performing one among the run champions -- top 5 run champions all have this design. While having diversity among the run champions could be considered a feature, it also suggests poor search in the design space, as the search algorithm fails to take advantage of this particular high-performance region of the search space~\cite{mertan2024investigating,cheney_difficulty_2016}. The \textit{Pollination} treatment more consistently finds this design in independent repetitions, indicating a more-global search.

\section{Discussion}\label{sect:discussion}

In previous sections, we hypothesized that promoting migrations would result in a better search for MAP-Elites algorithms in the context of brain-body co-optimization. We proposed the use of distilled controllers~\cite{mertan2024towards} for this purpose. Our results show that the proposed pollination strategy, replacing some individuals' controllers with the distilled controller, does increase the number of migrations and results in better total and maximal quality. However, injecting distilled controllers into the population not only increases the success of body mutations and increases the number of migrations but also could affect the success of brain mutations. To more conclusively argue that more migrations enabled by distilled controllers resulted in a better search, not increased brain power, we compare \textit{Standard} and \textit{Pollination} treatments in terms of how each type of mutation changes the fitness of offspring under treatments. 

\begin{figure}
    \centering
    \includegraphics[width=0.5\linewidth]{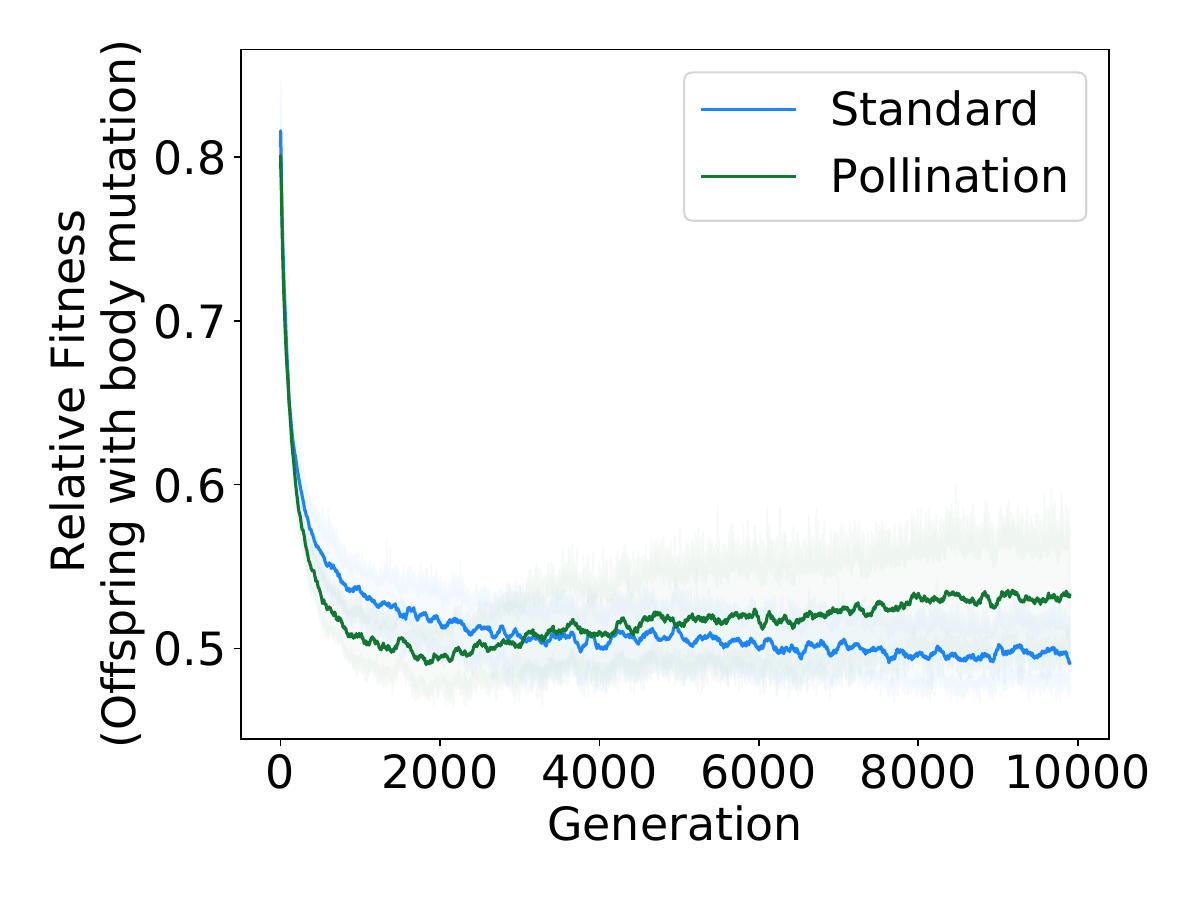}%
    \includegraphics[width=0.5\linewidth]{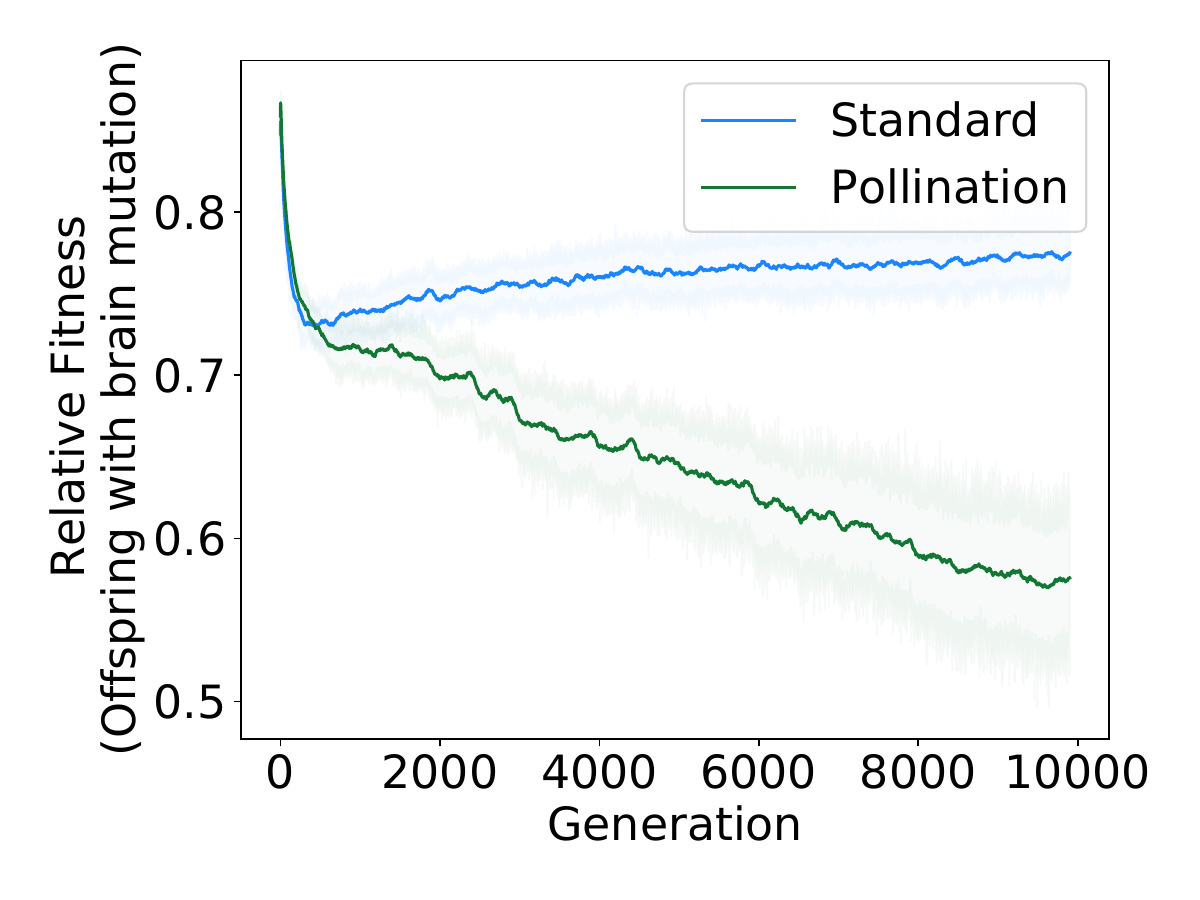}
    \caption{Fitness of offspring with body mutations (left) and brain mutations (right) relative to their parents' fitness throughout evolution. A moving average with a window length of 100 is applied to smooth the data to make the trends more apparent. While the \textit{Pollination} treatment helps with the success of body mutations compared to the \textit{Standard} treatment, it turns out distilled controllers are more fragile to brain mutations, decreasing the success of such mutations. }
    \label{fig:compare-std-pollination-offspring-fitness}
    \Description{desc}
\end{figure}

Fig.~\ref{fig:compare-std-pollination-offspring-fitness} shows the fitness of the offspring relative to their parents when they acquire body (Fig.~\ref{fig:compare-std-pollination-offspring-fitness} (left)) or brain (Fig.~\ref{fig:compare-std-pollination-offspring-fitness} (right)) mutations, averaged at each generation and over 20 repetitions. To make the trends in the data clear, we also applied a moving average with a window of length 100 to smooth the signal. Interestingly, \textit{Pollination} helps with body mutations but also makes controllers more susceptible to harmful brain mutations. This not only indicates that the increased performance is indeed a result of more migrations (as the search over the controller space becomes harder in \textit{Pollination} treatment) but also suggests a trade-off -- controllers that generalize better to other morphologies (that are more robust to body mutations) are harder to optimize.

Although we can find a trade-off that translates into better quality-diversity metrics at the end results, it is unlikely that we find the best trade-off. This trade-off invites further research into optimizing generalist controllers. Recent work~\cite{mertan2024towards,huang_one_2020,pathak_learning_2019,kurin_my_2021,gupta2022metamorph} focuses on how to model the architecture of such generalist controllers or how to train them, but there are not many investigations into the properties of such controllers. Enabling better brain-body co-optimization with quality-diversity algorithms such as MAP-Elites will require a better understanding of this trade-off.

\begin{figure}
    \centering
    \includegraphics[width=0.5\linewidth]{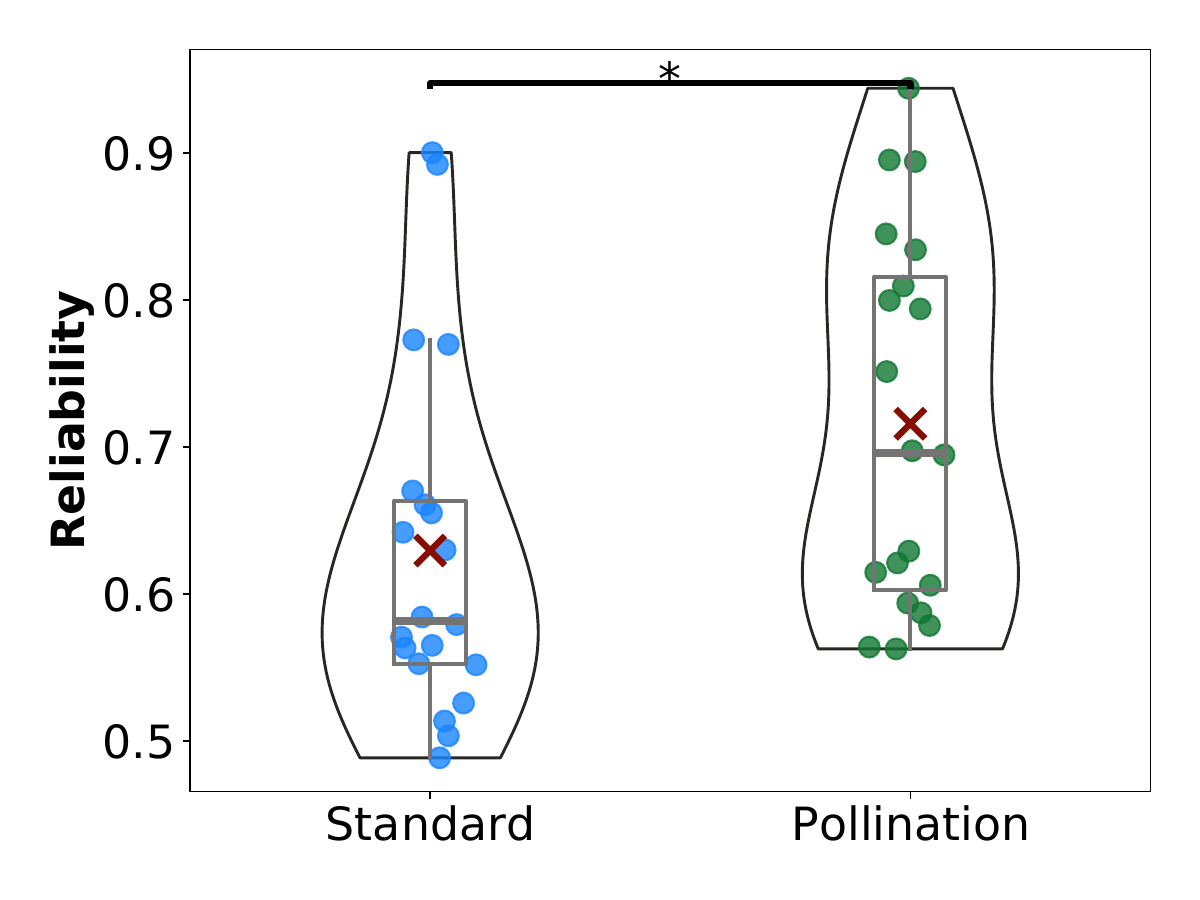}%
    \includegraphics[width=0.5\linewidth]{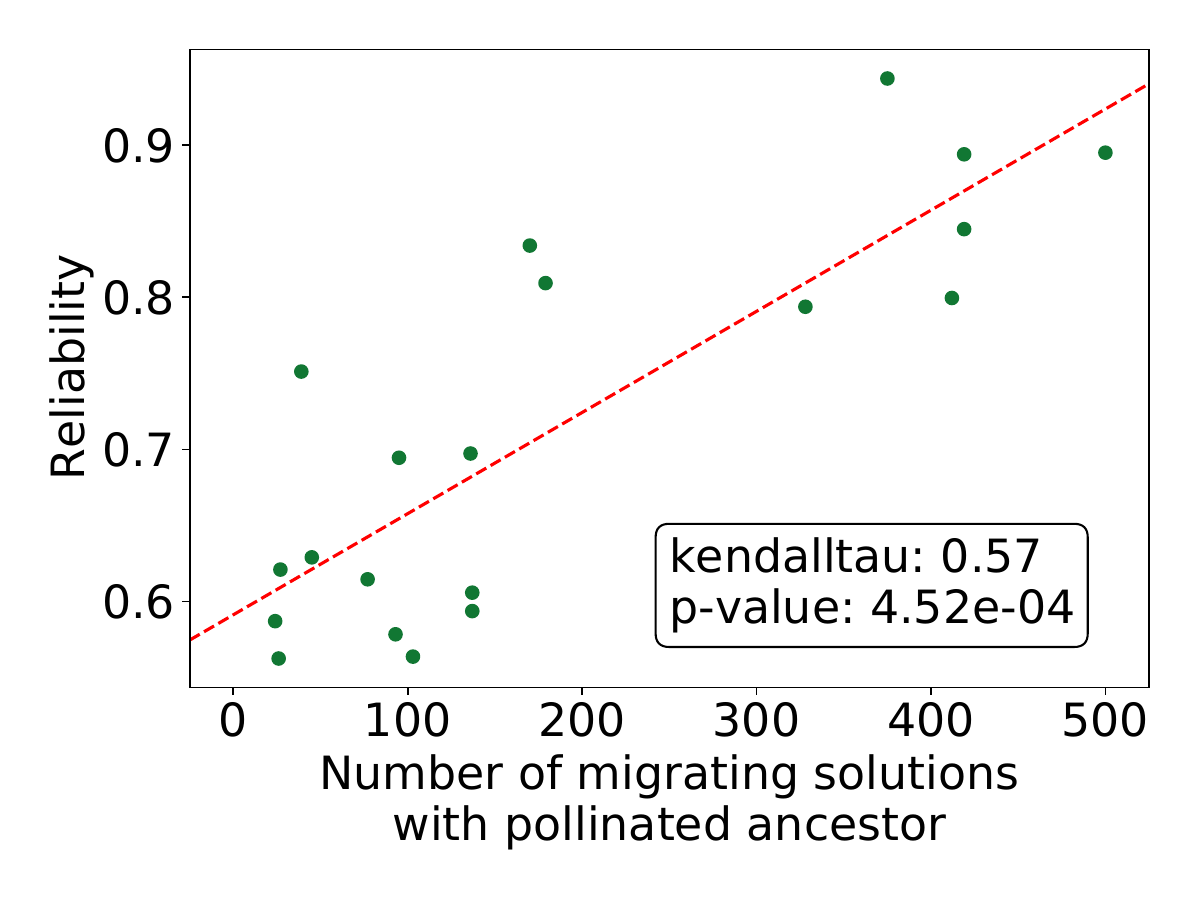}
    \caption{Reliability of the \textit{Standard} and \textit{Pollination} treatments (left) and pollination-induced migrations against reliability (right). Increased migrations that the \textit{Pollination} treatment enabled correlate strongly with reliability.}
    \label{fig:compare-std-pollination-reliability}
    \Description{desc}
\end{figure}

The findings we share in this work also have implications for quality-diversity algorithms like MAP-Elites. We show, in our test case of brain-body co-optimization, that the problem at hand may resist the creation of good stepping stones, and encouraging more migrations could alleviate this issue. While this seems like a straightforward idea, we believe it is valuable to point it out explicitly, as it has not been explicitly discussed to the best of our knowledge. For example, the original MAP-Elites work~\cite{mouret_illuminating_2015} shows lower opt-in reliability for the MAP-Elites algorithm when it is applied to soft robot morphology optimization (note that this problem is not brain-body co-optimization, but our point still applies). Opt-in reliability roughly measures an algorithm's ability to find the best solution\footnote{The best performance for each niche is approximated by pooling all the data from all runs and all treatments.} for each niche in a single run, if it finds a solution for that niche at all (hence opt-in)\footnote{In comparing \textit{Standard} and \textit{Pollination} treatments, opt-in reliability is equal to reliability, as both algorithms fill all the available niches in all repetitions.}. Measurement of the reliability of \textit{Standard} and \textit{Pollination} treatments also show an advantage for the \textit{Pollination} treatment (Fig.~\ref{fig:compare-std-pollination-reliability}, left) with a strong correlation with pollination-induced migrations (Fig.~\ref{fig:compare-std-pollination-reliability}, right). This suggests a relation between migrations and the reliability of the algorithm -- encouraging more migrations improves the reliability.

Lastly, we would like to connect our proposed methodology with a broader understanding of evolutionary algorithms. While natural evolution is capable of creating hierarchically organized complex structures, artificial evolution seems to lack this ability. It is not clear whether a fundamental piece is missing in our algorithms or whether it is a matter of scale, substrate, problem domain, etc. Despite the lack of understanding, we believe that it is worthwhile to seek more inspiration from natural evolution and how it creates complex structures. One theory that attempts to explicate how evolving systems form complex structures is Evolution Mechanics (EM)~\cite{sevinchan2021}. In summary, EM considers three fundamental processes that act on self-replicators (SR): (1) diversification is the process of SRs accumulating mutations that allow them to occupy accessible niches, (2) aggregation is the process of SRs forming loose collectives, and (3) transformation is the process of such collectives becoming new SRs by starting to reproduce together as a new whole -- evolutionary transition in individuality. At the end of this cycle, we end up with a "higher-level" SR that is more complex. Eventually, this cycle can repeat many times, giving rise to hierarchically organized complex structures that we see in nature.

We see recent approaches that distill the diverse knowledge in the archive (also referred to as archive distillation)~\cite{mertan2024towards,faldor2023map,mace2023quality} as conceptually similar to EM's aggregation process. Quality-diversity algorithms such as MAP-Elites can effectively fill accessible niches with high-performing solutions (diversification), and distilling from this archive can create more capable and complex solutions (aggregation). In our work, we take a step forward by injecting such solutions back into the population (transformation), closing the EM cycle for further evolution. 
Although the proposed method of distilling controllers is applied here in the context of brain-body co-optimization, we speculate that the idea of using a distillation-like process to aggregate diverse solutions and use them to pollinate a population/map for further evolution may hold the potential to instantiate and explore the EM framework in a wide variety of evolutionary algorithms and problem domains.

\section{Conclusion}

In this work, we investigated how the MAP-Elites~\cite{mouret_illuminating_2015} algorithm behaves when applied to the brain-body co-optimization problem of virtual voxel-based soft robots. We showed that the number of migrations throughout evolution decreases rapidly, and in some cases, migrations that occur in the lineage of run champions occur at a low fitness regime and very early on in the evolution. Moreover, we ran a control experiment where we do not allow migrations after $N$\textsuperscript{th} generation, demonstrating that the later migrations are not useful for maximal quality. These findings prompted us to investigate how the number of migrations can be increased to potentially improve quality-diversity metrics. Based on previous research investigating the dynamics of evolutionary brain-body co-optimization~\cite{mertan2024investigating,mertan_modular_2023,cheney_difficulty_2016}, we proposed a method, Pollination, where we periodically replace some solutions' controllers with an empowered controller~\cite{mertan2024towards}. We showed that this process increases the success of body mutations and enables more migrations, which in turn improves quality-diversity metrics such as total quality, maximal quality, and reliability.

\begin{acks}
This material is based upon work supported by the National Science Foundation under Grant No. 2008413, 2239691, 2218063.
Computations were performed on the Vermont Advanced Computing Core supported in part by NSF Award No. OAC-1827314.
\end{acks}

\bibliographystyle{ACM-Reference-Format}
\bibliography{sample-base}


\end{document}